\documentclass[acmsmall]{acmart}
\AtBeginDocument{%
  \providecommand\BibTeX{{%
    \normalfont B\kern-0.5em{\scshape i\kern-0.25em b}\kern-0.8em\TeX}}}

\setcopyright{acmcopyright}
\copyrightyear{2023}
\acmYear{2023}
\acmDOI{XXXXXXX.XXXXXXX}

\usepackage{amssymb}
\usepackage{amsmath}
\usepackage{algorithm} 
\usepackage{algorithmic}
\usepackage{natbib}
\usepackage{bbm}
\usepackage{bm}
\usepackage{setspace}
\usepackage{multirow}
\usepackage{hyperref}
\usepackage{tikz}
\usetikzlibrary{arrows,shapes,positioning}
\usetikzlibrary{calc,decorations.markings}
\acmJournal{JACM}

\begin{document}

\title{Transfer Learning for Bayesian Optimization: A Survey}

\author{Tianyi Bai}
\affiliation{%
  \streetaddress{School of Mathematics and Statistics}
  \institution{Beijing Institute of Technology}
  \city{Beijing}
  \country{China}
}
\email{baitianyi@bit.edu.cn}

\author{Yang Li}
\authornotemark[1]
\affiliation{%
  \streetaddress{Data Platform, TEG}
  \institution{Tencent Inc.}
  \city{Beijing}
  \country{China}}
\email{thomasyngli@tencent.com}

\author{Yu Shen, Xinyi Zhang}
\affiliation{%
  \streetaddress{Key Lab of High Confidence Software Technologies}
  \institution{Peking University}
  \city{Beijing}
  \country{China}
}
\email{{shenyu, zhang_xinyi}@pku.edu.cn}

\author{Wentao Zhang}
\affiliation{%
  \institution{Mila - Québec AI Institute}
  \city{Montréal}
  \country{Canada}
}
\email{wentao.zhang@mila.quebec}

\author{Bin Cui}
\authornote{Yang Li and Bin Cui are the corresponding authors.}
\affiliation{%
 \streetaddress{Key Lab of High Confidence Software Technologies} 
 \institution{Peking University}
 \city{Beijing}
 \streetaddress{Institute of Computational Social Science, Peking University}
 \city{Qingdao}
 \country{China}
}
\email{bin.cui@pku.edu.cn}

\renewcommand{\shortauthors}{Bai et al.}

\begin{abstract}
A wide spectrum of design and decision problems, including parameter tuning, A/B testing and drug design, intrinsically are instances of black-box optimization.
Bayesian optimization (BO) is a powerful tool that models and optimizes such expensive ``black-box'' functions.
However, at the beginning of optimization, vanilla Bayesian optimization methods often suffer from slow convergence issue due to inaccurate modeling based on few trials.
To address this issue, researchers in the
BO community propose to incorporate the spirit of transfer learning to accelerate optimization process, which could borrow strength from the past tasks (source tasks) to accelerate the current optimization problem
(target task).
This survey paper first summarizes transfer learning methods for Bayesian optimization from four perspectives: initial points design, search space design, surrogate model, and acquisition function.
Then it highlights its methodological aspects and technical details for each approach.
Finally, it showcases a wide range of
applications and proposes promising future directions.

\end{abstract}



\maketitle

\section{Introduction}

Black–box optimization (BBO) is the task of optimizing an objective function within a limited budget for function evaluations. 
``Black-box'' means that the objective function has no analytical form. 
In this way, we cannot access but we can only observe its outputs (i.e., objective values) based on the given inputs, without any knowledge of its internal workings. 
Since the evaluation of objective functions is often expensive, the goal of black-box optimization is to find the global optimum as rapidly as possible~\cite{li2021openbox}.

Black-box optimization problems appear everywhere. 
Design problems and many decision problems, which are pervasive in scientific and industrial endeavours, fall into the domain of black-box optimization, including experiment design~\cite{gardner2014bayesian,ueno2016combo,greenhill2020bayesian}, machine design~\cite{khurmi2005textbook,shigley2004standard}, drug design~\cite{imani2020bayesian,pyzer2018bayesian}, robotics~\cite{lizotte2007automatic,martinez2007active}, environmental monitoring~\cite{marchant2012bayesian,barrenetxea2008sensorscope}, combinatorial optimization~\cite{hutter2011sequential,kkorte2011combinatorial}, and automatic machine learning~\cite{bergstra2011algorithms,feurer2015efficient,li2020efficient,li2022volcanoml,erickson2020autogluon,jin2019auto}, etc. 

\paragraph{Example 1: Hyperparameter tuning}
The performance of machine learning (ML)
models heavily depends on the choice of hyperparameter configurations (e.g., regularization parameter in support vector machine or learning rate in a deep neural network). 
As a result, automatically tuning the hyperparameters has attracted lots of interest in machine learning community. 

\paragraph{Example 2: A/B Testing}
A/B testing is useful for understanding user engagement and satisfaction of online features like a new feature or product. 
Large social media sites like LinkedIn, Facebook, and Instagram use A/B testing to make user experiences more successful and as a way to streamline their services~\cite{xu2015infrastructure}.
A/B testing is widely used by data engineers, designers, software engineers, and entrepreneurs, among others.
For instance, A/B testing can be utilized to determine the most suitable price for the product, where it aims to find out which price-point maximizes the total revenue.

\paragraph{Example 3: Knobs tuning}
Modern database management systems (DBMS) contain tens to
hundreds of critical performance tuning knobs that determine the
system runtime behaviors.
Different knobs directly affect the running database performance in terms of latency and throughput.
Recently, many methods are proposed to utilize ML based techniques to optimize the performance of DBMSs automatically.

\paragraph{Example 4: Big data platforms tuning}
Spark has emerged as one of the most widely used frameworks for massively parallel data analytics. 
Spark task is controlled by up to 160 configuration parameters, which determine many aspects including dynamic
allocation, scheduling, memory
management, execution behavior, etc. 
Tuning arbitrary Spark applications by efficiently and automatically navigating over the huge search space is a challenging task.

\paragraph{Example 5: Electronic design automation}
Electronic design automation (EDA) tools play a vital role
in pushing forward the VLSI industry. 
The design complexity keeps increasing in order to ensure timing, reliability, manufacturability, etc.
This trend brings the increasing amount of parameters involved in EDA tools, thus incurring a huge design search space.
The aim is to find the most suitable parameters in EDA tools to achieve desired quality~\cite{ma2019cad}.

Recently, Bayesian Optimization (BO) methods have become one of the most prevailing frameworks in solving black-box optimization problems~\cite{shahriari2015taking}.
BO-based solutions have been extensively investigated and deployed to solve the BBO problems efficiently and effectively, including the aforementioned examples~\cite{hutter2011sequential,bergstra2011algorithms,snoek2012practical,li2021mfes,hypertune,zhang2021facilitating,zhang2022towards,alipourfard2017cherrypick,ma2019cad}.

\paragraph{Challenge}
Although Bayesian optimization (BO) methods have achieved a great stride of success in a wide range of fields, there still remain issues that need to be addressed.
One of them is about the {\it slow convergence} issue, which greatly hampers the efficiency and practicality of BO. 
The main idea of BO is to use a surrogate model, typically a Gaussian Process (GP), to describe the relationship between a configuration and its performance, and then utilize this surrogate to determine
the next configuration to evaluate by optimizing an acquisition function that balances exploration and exploitation.
However, evaluating the objective functions is usually computationally expensive. 
Given a limited budget, few observations about the function evaluations are obtained, and these observations cannot be used to learn an accurate surrogate model that represents the objective function well.
Further, the surrogate model cannot guide the search of configuration effectively and efficiently, thus leading to the ``slow convergence'' problem.
In many real scenarios, users cannot bear the additional cost for initial trials during the cold start period.
For each trial, the cost in terms of device, expense or time can be very expensive.
In addition, along with the growing search space, the number of trials increases for building accurate surrogates.
Therefore, it is essential to improve Bayesian optimization method with faster convergence.

\paragraph{Opportunity}
To address this issue, researchers in the BO community propose to incorporate the spirit of transfer learning to accelerate Black-box optimization, which could borrow strength from past tasks (source tasks) to accelerate the current optimization task (target task). 
Many real-world black-box problems usually need to be constantly re-optimized as task/environment changes, e.g., the update of model/code in the AutoML applications. The optimal configuration (i.e., some design or decision) may also change as the task/ environment varies, and so should be frequently re-optimized. Although they may change significantly, the region of good or bad configurations may still share some correlation with those of previous tasks, and this provides the opportunity for faster Bayesian optimization.

\paragraph{Method categorization}
In this paper, we review the transfer learning methods for Bayesian optimization in depth.
The overview of the categorization is summarized in Table~\ref{table1}.
As far as we know, Bayesian Optimization consists of four main components that can be customized manually, which are the initial points, the search space, the surrogate model, and the acquisition function. 
Based on this perspective, we divide existing transfer learning methods into four main categories. 
For each main category, we further divide each category based on specific techniques.

\begin{table}[h]
\caption{Transfer learning for Bayesian Optimization: Overview}
\label{table1}
\scalebox{0.95}[.95] {
\begin{tabular}{|c|c|}
\hline
\textbf{BO components} & \textbf{Specific categories} \\ \hline
\multirow{3}{*}{\textit{\begin{tabular}[c]{@{}c@{}}Surrogate\\ Design\end{tabular}}} & \begin{tabular}[c]{@{}c@{}}Gaussian Process as surrogate model (Kernel Design, \\ Prior Design, Data Scale Design, Ensemble Design)\end{tabular} \\ \cline{2-2} 
 & Bayesian Neural Network as surrogate model \\ \cline{2-2} 
 & Neural Process as surrogate model \\ \hline
\multirow{3}{*}{\textit{\begin{tabular}[c]{@{}c@{}}Acquisition function\\ Design\end{tabular}}} & Multi-task BO acquisition function \\ \cline{2-2} 
 & Ensemble GPs-based acquisition function transfer \\ \cline{2-2} 
 & Reinforcement learning-based acquisition function transfer \\ \hline
\multirow{3}{*}{\textit{\begin{tabular}[c]{@{}c@{}}Initialization \\Design\end{tabular}}} & Meta-features based initialization \\ \cline{2-2} 
 & Gradient-based learning initialization \\ \cline{2-2} 
 & Evolutionary algorithm based initialization \\ \hline
\multirow{2}{*}{\textit{\begin{tabular}[c]{@{}c@{}}Search space \\ Design\end{tabular}}} & Search space pruning method \\ \cline{2-2} 
 & Promising search space design \\ \hline
\end{tabular}
}
\end{table}

\paragraph{Contribution and Overview}
In this survey, the main contributions can be summarized as follows:
\begin{enumerate}
    \item We systematically categorize existing transfer learning works of Bayesian optimization based on ``what to transfer'' and ``how to transfer''. Problem setups are from the ``what'' perspective, indicating which learning process we want to make transfer. Techniques are from the ``how'' perspective, introducing the methods proposed to solve BO problems. For each category, we present detailed method descriptions for reference.
    \item We propose and discuss a general transfer learning framework for Bayesian optimization. Such a framework can act as a guidance for developing new approaches.
    \item In addition, we also present the potential application scenarios, where the transfer learning approaches for Bayesian optimization could work well.
\end{enumerate}

We begin in Section~\ref{sec_background}, with an introduction to black-box optimization and Bayesian optimization.
In Sections~\ref{sec_overview}-\ref{sec_space}, we introduce existing transfer learning methods from four aspects, Surrogate Design in Section~\ref{sec_surrogate}, Acquisition function Design in section~\ref{sec_af}, Warm-start Method in Section~\ref{sec_warmstart}, and Search space Design in Section~\ref{sec_space}.
We provide the description for potential application scenarios in Section~\ref{sec_apps}, and end this survey with a conclusion in Section~\ref{sec_conclusion}.

\section{Background and Formulation}
\label{sec_background}
\subsection{Black-box Optimization}
{\em Black-box Optimization} (BBO) is a kind of optimization problem when the objective function is a black-box function. On the contrary to white-box function, black-box function has no exact form and is not access to any other information like gradients or the Hessian. The mathematical expression of black-box function is $f:\mathcal{X}\to\mathcal{R}$, where $\mathcal{X}$ is the search space for a certain problem. For a given point $x\in\mathcal{X}$, we can evaluate the function value $f(x)$ of a black-box function. When the evaluation cost is very high, selecting which point to evaluate next becomes a vital problem to consider. Therefore, the problem of BBO can be interpreted as to approach the global optimum as rapidly as possible through selecting a sequence of search points $\{x_t\}_{t=1}^{n}$ and evaluating their function value.

Black-box Optimization has a wide application in many areas where the relationship of inputs and outputs is complex or unknown, such as automated hyperparameters tuning of automated machine learning system, optimization of chemical compounds or materials~\cite{terayama2021black}, reference learning and interactive interfaces~\cite{brochu2010bayesian}, resource allocation and so on. 

Due to the lack of information of the target function, in order to solve a BBO problem, we have to utilize some navigation algorithms to guide our searching process. There exists two main taxonomies for those BBO algorithms, which could be summarized as non-adaptive or self-adaptive algorithms, local optimization or global optimization algorithms. The simplest one is algorithms with no adaptive capacity, including {\em Grid Search} that selects $x_t$ along a grid made of Cartesian product of all candidates values, and {\em Random Search} that selects $x_t$ uniformly at random from $\mathcal{X}$ at each steps. The self-adaptive algorithms consists of classic algorithms (such as {\em Simulated Annealing}), population-based optimization algorithms~\cite{xiao2015optimization} (such as {\em Genetic Algorithms}~\cite{doerr2015black}, {\em Ant Colony Optimization}~\cite{dorigo2019ant}) and so on. As for local or global optimization algorithms, the main difference between them is that local optimization algorithms can only get a local optimum, but global optimization algorithms try their best to get a global optimum. Many local optimization algorithms try to maintain simple models of the objective function $f$ within a subset of the feasible regions (known as trust region), including derivative-free optimization~\cite{conn2009introduction} (such as {\em Nelder-Mead simplex reflection}~\cite{nelder1965simplex}). While the global optimization algorithms try to optimize the function in the overall searching spaces to obtain a global optimum.

More recently, {\em Bayesian Optimization} has been developed to solve BBO problem~\cite{mockus1978application} and is been shown to outperform other global optimization algorithms on a number of challenging optimization benchmark functions\cite{jones2001taxonomy}. Bayesian optimization utilizes the idea from multi-armed bandit problems to manage exploration and exploitation trade-offs. This optimization technique goes under a Bayesian pattern, which learns a posterior from a given prior and the observed information of sequential evaluation.

For a certain BBO problem, there are three main questions to consider, the design of search space, the selection of navigation algorithm and initialization. Due to the high computational complexity of searching the whole flexible region in large data-sets, many researchers have come up with an idea of removing unpromising region to accelerate searching process
~\cite{wistuba2015hyperparameter,perrone2019learning}. Besides, sometimes we can acquire little information from the certain problem, thus designing a bounded search space may be hard to accomplish. Therefore, some works propose to incrementally expand the search space with unbounded form~\cite{shahriari2016unbounded,nguyen2019filtering}. It should be noted that there is a difference between traditional local optimization algorithms and the algorithms with search space design, as the latter ones still belong to the global optimization category that hopes to find a global optimum. Meanwhile, as many navigation algorithms are self-adaptive algorithms, choosing a promising initial point is also beneficial for further searching process~\cite{kazimipour2014review,feurer2015initializing}.


\subsection{Bayesian Optimization}
{\em Bayesian Optimization} (BO)~\cite{mockus1978application} is one of the state-of-the-art algorithms for black-box optimization. It performs well when expensive function is needed to be evaluated and it has applied in many areas~\cite{shahriari2015taking}, including robotics~\cite{berkenkamp2021bayesian}, automatic machine learning~\cite{snoek2012practical,klein2017fast,li2021mfes}, environmental monitoring~\cite{marchant2012bayesian}, reinforcement learning~\cite{brochu2010tutorial}, neural architecture search~\cite{kandasamy2018neural}and so on.

The main problem for Bayesian optimization to solve is mathematically as follow. We consider the problem as an optimization problem for an unknown objective function $f$, and we hope to find a global minimizer (or maximizer) of that function:
\begin{equation}
    \bm{x^*}=\mathop{\arg\min}_{\bm{x} \in \mathcal{X}}f(\bm{x}),
    \label{eq:original}
\end{equation}
where $\mathcal{X}$ is a designed search space. Bayesian optimization can not only deal with the traditional problem that the search space is numerical and in a form of $\mathbb{R}^d$, but it can also applied to problems with unusual search spaces, including categorical or conditional inputs, or even combinatorial search spaces with multiple categorical inputs. Moreover, the black-box function $f$ is assumed with no simple closed form, but it can be evaluated at any point in the search space. For any given point $x_i$, we can get an noisy observation of function $f$, noted as $y_i=f(x_i)+\varepsilon$ in which $\varepsilon\sim\mathcal{N}(0,\sigma^2)$ and $\mathbb{E}[y\mid\bm{x}]=f(\bm{x})$.

In this setting, we consider Bayesian optimization as a sequential search algorithm. At iteration n, BO uses the evaluated information to guide the search of new location $\bm{x}_{n+1}$ and get the noisy evaluation $\bm{y}_{n+1}$ of the black-box function $f$. And after N rounds of iterations, BO makes a final decision of the optimization solution and provide a solution noted as $\bm{\hat{x}^*}$.

Bayesian optimization mainly contains two key ingredients, a probabilistic surrogate model and an acquisition function. We assume that the black-box function $f$ is sampled from a probabilistic distribution, known as probabilistic surrogate model, which contains our beliefs on the unknown black-box function and captures the new observation information to update our knowledge of the current function. 
The acquisition function is used to balance the exploration and exploitation trade-off to make decision of next searching point in the domain. We will introduce common surrogate models and acquisition function in the following parts.
\begin{algorithm}[tb]
  \small
  \caption{Pseudo code for Bayesian Optimization}
  \label{algo:bo}
  \begin{algorithmic}[1]
  \REQUIRE the number of trials $T$, the hyper-parameter space $\mathcal{X}$,  surrogate model $M$, acquisition function $\alpha$, and initial hyper-parameter configurations $X_{init}$.
  \FOR{\{$\bm{x} \in X_{init}\}$}
    \STATE evaluate the configuration $\bm{x}$ and obtain its performance $y$.
    \STATE augment $D = D \cup (\bm{x}, y)$.
  \ENDFOR
  \STATE initialize observations $D$ with initial design.
  \FOR{\{ $i = |X_{init}| + 1, ..., T\}$}
  \STATE fit surrogate $M$ based on observations $D$.
  \STATE select the configuration to evaluate: $\bm{x}_i=\operatorname{argmax}_{\bm{x}\in \hat{\mathcal{X}}}\alpha(\bm{x}, M)$.
  \STATE evaluate the configuration $\bm{x}_i$ and obtain its performance $y_i$.
  \STATE augment $D=D\cup(\bm{x}_i,y_i)$.
  \ENDFOR
  \STATE \textbf{return} the configuration with the best observed performance.
\end{algorithmic}
\end{algorithm}

\subsubsection{Surrogate model}
\label{surrogate model}
There are many available surrogate models for Bayesian Optimization. Most of the researches use Gaussian Processes~\cite{snoek2012practical}, Bayesian neural networks~\cite{snoek2015scalable,perrone2018scalable,springenberg2016bayesian}, tree parzen estimators~\cite{bergstra2011algorithms}, or random forest~\cite{breiman2001random,hutter2011sequential} as surrogate models. In this section, we will introduce some of them, and the detailed settings will be introduced in following sections.

\noindent
\textbf{$\bullet$ Gaussian Processes as surrogate models.}
{\em Gaussian Processes} (GPs)~\cite{williams2006gaussian} has been widely used as surrogate models in Bayesian Optimization~\cite{mockus1978application,hutter2011sequential,snoek2012practical}. The Gaussian process has a convenient property that, if we assume prior as a Gaussian distribution, we can get the posterior by computing the mean and covariance function, which still follows a Gaussian distribution.

Usually we assume the objective black-box function follows a Gaussian distribution prior. Besides, we assume that each variable $f_i=f({\bm x}_i)$ is independent and identically distributed to others, thus the joint distribution of $\bm{f}:=f_{1:n}$ is a joint Gaussian prior, i.e. $\bm{f}\mid\bm{X}\sim \mathcal{N}(\bm{m},\bm{K})$, where $\bm{X}$ is a vector consists of $\{{\bm x}_i\}_{i=1}^n$, $\bm{m}$ is a mean vector consists of mean values $\{{m_i=\mu_0({\bm x}_i)}\}_{i=1}^n$ generated from mean function $\mu_0:\mathcal{X}\to\mathcal{R}$, and $\bm{K}$ is a covariance matrix consists of positive–definite kernels ${\{K_{i,j}=k({\bm x}_i,{\bm x}_j)}\}_{i,j=1}^n$ generated from convariance function $k:\mathcal{X}\times\mathcal{X}\to\mathcal{R}$. The noisy observations $\bm{y}:= y_{1:n}$ naturally follow a normal distribution given as,
\begin{equation}
    \bm{y}\mid\bm{f},\sigma^2\sim \mathcal{N}(\bm{f},\sigma^2\bm{I}).
    \label{eq:noisy}
\end{equation}

Given a set of observation $\mathcal{D}=\{{\bm x}_i,y_i\}_{i=1}^n$, also noted as $\mathcal{D}=(\bm{X},\bm{y})$, we can compute the posterior distribution of $f$ at an arbitrary test point $\bm x$ by computing its posterior mean and variance function:
\begin{equation}
\begin{split}
    &f\mid {\bm x},\bm{X},\bm{y}\sim \mathcal{N}(\mu_n({\bm x}),\sigma^2_n({\bm x})), where\\
    \mu_n({\bm x})&=\mu_0({\bm x})+k(\bm{X},\bm{x})^T(\bm{K}+\sigma^2\bm{I})^{-1}(\bm{y}-\bm{m}),\\
    \sigma^2_n({\bm x})&=k({\bm x},{\bm x})-k(\bm{X},{\bm x})^T(\bm{K}+\sigma^2\bm{I})^{-1}k(\bm{X},{\bm x}),
    \label{eq:posterior}
\end{split}
\end{equation}
where $k(\bm{X},{\bm x})$ is a vector that shows the result of computing covariance function between ${\bm x}$ and $\bm{X}$. 

Usually we require a predetermined form of the mean function $\mu_0$ and covariance function $k$. Previous works usually set mean functions to be zero or linear, and the popular kernel functions include {\em Mat\v{e}rn kernels}, {\em Squared Exponential kernel} and {\em RBF kernel}~\cite{rasmussen2003gaussian,muandet2017kernel}. The hyper-parameters in these functions are usually trained by maximizing data-likelihood of the current observations, or by putting a prior on the mean/kernel hyper-parameters and obtaining a distribution of such hyper-parameters to adapt the model given observations \cite{rasmussen2003gaussian}.


\noindent
\textbf{$\bullet$ Random Forests as surrogate models. }
{\em Random Forests} are an ensemble of regression trees that are used to handle the problems with many input variables and hard to be dealt with a single regression tree. Regression trees~\cite{breiman2017classification} utilize tree structures to model classification or regression problems in machine learning. Different from typical decision trees that also leverage tree structures, regression trees have real values rather than classifying labels at their leaves, thus they can give out a predictive value for every input point. It is proved that the random forest method always converges to the optimal solutions~\cite{breiman2001random} and empirically performs well especially on problems with categorical inputs.

Previous works have utilized random forests in Bayesian optimization (such as {\em Sequential Model-based Algorithm Configuration} (SMAC) in~\cite{hutter2011sequential}), due to their efficiency when dealing with categorical inputs, and their advantage that can give out both predictive value and uncertainty of the prediction for any given input. To construct a random forest, independent regression trees are built by randomly sampling $n'$ points from a given dataset $\mathcal{D}=\{{\bm x}_i,y_i\}_{i=1}^n$, and then randomly selecting features to split the points in every node. Assuming a random forest has $m$ regression trees in it, we note $T_i$ as the predictive function of the $i$-th regression tree. The total predictive mean is given as the average of predictive values from each regression tree in the random forest, as $\mu({\bm x})=\frac{1}{m}\sum_{i=1}^m T_i({\bm x})$, and the variance is given as $\sigma^2({\bm x})=\frac{1}{m-1}\sum_{i=1}^m(T_i({\bm x})-\mu({\bm x}))^2$. 

\noindent
\textbf{$\bullet$ Tree Parzen Estimators as surrogate models. }
While GPs model $p(y\mid {\bm x})$ directly, {\em Tree Parzen Estimators}~\cite{bergstra2011algorithms} model $p({\bm x}\mid y)$ and $p(y)$ separately. Specifically, to model $p({\bm x}\mid y)$, a parameter $\gamma$ that tells the selected quantile should be given, thus for a given dataset $\mathcal{D}=\{{\bm x}_i,y_i\}_{i=1}^n$, $\gamma$ can be used to choose the observation value $y'$ that satisfies $p(y<y')=\gamma$. Leveraging this chosen observation value $y'$, the 
likelihood $p({\bm x}\mid y)$ can be defined as
\begin{equation}
    p({\bm x}\mid y)=
    \begin{cases}
    l({\bm x}) &\  \text{if}~y<y'\\
    g({\bm x})&\  \text{if}~y\geq y'
    \end{cases}
\end{equation}
where $l({\bm x})$ is the probability computed by using the points in dataset $\mathcal{D}$ that satisfies $y({\bm x}_i)<y'$, and $g({\bm x})$ is the probability computed by using the rest of the points in dataset $\mathcal{D}$. Therefore, using Bayes rule, the posterior $p(y\mid {\bm x})$ can be given as 
\begin{equation}
    p(y\mid {\bm x})=\frac{p({\bm x}\mid y)p(y)}{p({\bm x})}
\end{equation}
where $p({\bm x})=\int_{\mathbb{R}}p({\bm x}\mid y)p(y)dy=\gamma l({\bm x})+(1-\gamma)g({\bm x})$. To utilizing this method in Bayesian optimization, previous works~\cite{bergstra2011algorithms,zoller2019survey} consider to combine it with the acquisition function Expected Improvement (general form of EI will be introduced in {Sec.\ref{af}}) as
\begin{equation}
\begin{split}
    \alpha_{EI_{y'}}({\bm x})&=\int_{-\infty}^{y'}(y'-y)p(y\mid {\bm x})dy=\int_{-\infty}^{y'}(y'-y)\frac{p({\bm x}\mid y)p(y)}{p({\bm x})}dy\\
    &=\frac{\gamma y'l({\bm x})-l({\bm x})\int_{-\infty}^{y'}p(y)dy}{\gamma l({\bm x})+(1-\gamma)g({\bm x})}\propto\frac{l({\bm x})}{\gamma l({\bm x})+(1-\gamma)g({\bm x})}
\end{split}
\end{equation}

\noindent
\textbf{$\bullet$ Bayesian Neural Networks as surrogate models. }
From the computational formula of GP posterior in {Eq.\ref{eq:posterior}} we can know that the inference time of a GP scales cubically with the number of observations, as it has to compute a dense covariance matrix and its inversion. For this reason, GP-based BO can hardly leverage large numbers of past function evaluations. Thus, for optimization problem that requires many evaluations, GP-based BO shows its weakness.

Therefore, some researchers have proposed to use {\em Bayesian Neural Network} as an alternative to GP to be the surrogate model of BO~\cite{snoek2015scalable,perrone2018scalable,springenberg2016bayesian}. Those works utilize the flexibility and scalability of neural networks while keep the well-calibrated uncertainty estimates of GPs. Those works use different methods to compute prior and posterior distribution, which we will introduce in detail in section 4.

\subsubsection{Acquisition function}
\label{af}
In Bayesian Optimization, acquisition functions are used to choose which point to evaluate next from the search space, i.e., based on the posterior model generated from the evaluated sets $\mathcal{D}_n$ to select the next querying point ${\bm x}_{n+1}$ in the search space $\mathcal{X}$. The main question an acquisition function has to handle is how to leverage the posterior model to manage the exploration and exploitation trade-offs. Usually, given an acquisition function $\alpha$, the next querying point ${\bm x}_{n+1}$ is computed by calculating the acquisition function for every point in a given search space and find the maximizer of it, i.e. 
\begin{equation}
    {\bm x}_{n+1}=\mathop{\arg\max}\limits_{{\bm x}\in \mathcal{X}}\alpha_n({\bm x})
\end{equation}

In general, acquisition functions can be divided into four categories (proposed by \citet{shahriari2015taking}), improvement-based policies (such as {\em Probability of Improvement}~\cite{kushner1964new}, {\em Expected Improvement}~\cite{movckus1975bayesian,jones1998efficient} and {\em Knowledge Gradient}~\cite{frazier2009knowledge}), optimistic policies (such as {\em Gaussian Process Upper Confidence Bound}~\cite{srinivas2009gaussian}), information-based policies (such as {\em Thompson Sampling}~\cite{thompson1933likelihood} and {\em Entropy Search}~\cite{hennig2012entropy}) and some portfolios containing multiple acquisition functions (such as {\em Entropy Search Portfolio}~\cite{shahriari2014entropy}). We will introduce some of them in the following.

In this section, we note $\mathcal{D}_n=\{{\bm x}_i,y_i\}_{i=1}^t$ as the dataset observed at $n$-th iteration, $f^*_n$ as the optimal evaluations at $n$-th iteration, i.e. $f^*_n=\mathop{\max}_{{\bm x}\in D_n}f({\bm x})$, and ${\bm x}^*=\mathop{\arg\max}_{{\bm x}\in D_n}f({\bm x})$.

\noindent
\textbf{$\bullet$ Probability of Improvement}
{\em Probability of Improvement} (PI) is an early method proposed by~\citet{kushner1964new} to manage the exploration-exploitation trade-offs. It simply utilizes the mean and variance function given by the probability model, and is defined as
\begin{equation}
    \alpha_{PI_n}({\bm x})=\mathbb{P}(f({\bm x})>f^*_n)=\Phi\left(\frac{\mu_n({\bm x})-f^*_n}{\sigma_n({\bm x})}\right),
\end{equation}
where the standard normal cumulative distribution function $\Phi$ is utilized to compute the cumulative probability.

\noindent
\textbf{$\bullet$ Expected Improvement}
{\em Expected Improvement} (EI) is proposed by~\citet{movckus1975bayesian} and popularized by~\citet{jones1998efficient}. It improves the PI as it considers the amount of improvement, thus not simply rely on the probability. The simplest form of EI can be note as
\begin{equation}
\begin{split}
    \alpha_{EI_n}({\bm x})&=\mathbb{E}_n[(f({\bm x})-f^*_n)\mathbb{P}(f({\bm x})>f^*_n)]\\
    &=(\mu_n({\bm x})-f^*_n)\Phi\left(\frac{\mu_n({\bm x})-f^*_n}{\sigma_n({\bm x})}\right)+\sigma_n({\bm x})\phi\left(\frac{\mu_n({\bm x})-f^*_n}{\sigma_n({\bm x})}\right)\\
    &=\sigma_n({\bm x})[\gamma({\bm x})\Phi(\gamma({\bm x}))+\phi(\gamma({\bm x}))],
\end{split}
\end{equation}
where $\phi$ is the probability density function of the standard normal distribution, and $\gamma({\bm x})=\frac{\mu_n({\bm x})-f^*_n}{\sigma_n({\bm x})}$.

\noindent
\textbf{$\bullet$ Gaussian Process Upper Confidence Bound}
{\em Gaussian Process Upper Confidence Bound} (GP-UCB) is proposed by~\citet{srinivas2009gaussian}. It is a method generated from the idea of using Gaussian Processes as surrogate models, and the GP-UCB is simply defined by utilizing the mean and variance function computed from the probability model, as
\begin{equation}
    \alpha_{GP-UCB_n}({\bm x})=\mu_n({\bm x})+\beta_n\sigma_n({\bm x}),
\end{equation}
where $\beta_n$ is a given parameter that control the degree of exploration and exploitation. Additionally, the {\em Gaussian Process Lower Confidence Bound} (GP-LCB) can be defined accordingly,
\begin{equation}
    lcb_n({\bm x})=\mu_n({\bm x})-\beta_n\sigma_n({\bm x}),
\end{equation}
which is also useful in some literature.

\noindent
\textbf{$\bullet$ Entropy Search}
{\em Entropy search} (ES) is proposed by \citet{hennig2012entropy} and it leverages the idea from information theory. Specifically, ES measures how promising a given point $\bm x$ is by computing the information gain of selecting it as next point to explore:
\begin{equation}
    \alpha_{ES_n}({\bm x})=H({\bm x}^*\mid D_n)-\mathbb{E}_{f({\bm x})}[H({\bm x}^*\mid D_n\cup \{({\bm x},\mu_n({\bm x}))\})]
    \label{es}
\end{equation}
where $H({\bm x}^*\mid D_n)$ represents the entropy of the posterior distribution $p({\bm x}^*\mid D_n)$ at $n$-th iteration. The evaluation at point ${\bm x}$ is approximated by utilizing the mean function $\mu_n$ computed from the probability model. And the expectation is taken over the posterior $f({\bm x})$, also given by the probability model. An alternative form is given as 
\begin{equation}
    \alpha_{ES_n}({\bm x})=\int\int[H({\bm x}^*\mid D_n)-H({\bm x}^*\mid D_n\cup \{({\bm x},\mu_n({\bm x}))\})]p(y\mid f)p(f\mid {\bm x})dydf
    \label{es-al}
\end{equation}
where it is usually approximated through sampling $f$ using Monte Carlo method due to the fact that it has no simple form. This form of ES is also applied in many works~\cite{swersky2013multi}. 

Some variant of ES has been proposed, such as {\em Predictive Entropy Search} (PES)~\cite{hernandez2014predictive} and {\em Max-value Entropy Search} (MES)~\cite{wang2017max}.

\subsection{Transfer Learning Scenarios}

We will give out a unified setting and notation for this problem.

\subsubsection{Settings and Notations}
\label{notations}

The primary notations used in this paper are listed in \textbf{Table \ref{tb:notation}}.

\begin{table}[!htbp]
\caption{Notations}
\vspace{1pt}
\centering
\begin{spacing}{1.2}
\begin{tabular}{p{3cm}p{8cm}}
\hline
	\multicolumn{1}{c}{\centering Symbol}
	&\multicolumn{1}{c}{\centering Definition}\\
\hline
	\centering$t_1, t_2,..., t_k$ &the source tasks\\
	\centering$t_T$ &the target task\\
	\centering$D^{t_1}_{n_1}$, ..., $D^{t_K}_{n_K}$ &the training history from $K$ source tasks\\
	\centering$D^{t_T}_{t}$ &the observations in the target task at $t$-th iteration\\
	\centering$f^i$ &the black-box function of task $t_i$\\
	\centering$\mathcal{X}^i$ &the search space of task $t_i$\\
	\centering$\bm y^{t_i}$ &the noisy evaluations of task $t_i$\\
 	\centering$\mu^i$ &the mean function of task $t_i$\\
   	\centering$\lambda^i$ &the variance function of task $t_i$, same as $\sigma^2$\\
   	\centering$\sigma^i$ &the standard deviation function of task $t_i$\\
 	\centering$k^i$ &the co-variance function of task $t_i$\\
   	\centering${\bm m}^i$ &the meta-features that extract the feature of datasets for task $t_i$\\
   	\centering${\bm x}^{t_i}$ &a given point of task $t_i$\\
	

\hline
\end{tabular}\label{tb:notation}
\end{spacing}
\end{table}

To avoid confusion of transfer learning problem and its notations, we propose a unified setting and notation for it.

We consider $K$ source tasks, noted as $t_1,t_2,...,t_K$, and one target task, noted as $t_T$. Our observations are taken from these $K+1$ tasks as input, in which $D^{t_1}_{n_1}$, ..., $D^{t_K}_{n_K}$ are training history from $K$ source tasks and $D^{t_T}_{t}$ is the observations in the target task at $t$-th iteration. The source task $t_i$ contains $n^i$ evaluated points $D^{t_i}_{n_i}=\{(\bm{x}_j^{t_i}, y_j^{t_i})\}_{j=1}^{n^i}$. Unlike $\{D^{t_i}_{n_i}\}_{i=1}^K$ that are obtained in previous tuning procedures, the number of observations $t$ in $D^{t_T}_{t}$ grows along with the current training process.

Given a task $t_i$, we note its black-box function as $f^i(x)$ and its search space as $\mathcal{X}^i$. The noisy evaluations $\bm y^{t_i}$ of task $t_i$ follow a distribution as {Eq.\ref{eq:noisy}} shows, which we note as $\bm{y}^{t_i}\sim \mathcal{N}(\bm{f}^{t_i},\lambda^{t_i}\bm{I})$ in transfer learning scenarios. We note $\mu^1,..., \mu^K,\mu^T$ as mean functions,  $k^1,...,k^K,k^T$ as co-variance functions, and $\lambda^1,...,\lambda^K,\lambda^T$ as the posterior variance functions of each task. Given a data-set on task $t^i$, noted as $D_{n_i}^{t_i}=\{(\bm{x}_j^{t_i}, y_j^{t_i})\}_{j=1}^{n^i}$, the posterior of task on a given point ${\bm x}^{t_i}$ can be noted as $f^i\mid{\bm x}^{t_i},{\bm X}^{t_i},{\bm y}^{t_i}\sim \mathcal{N}(\mu_{n_i}^i({\bm x^{t_i}}),\lambda_{n_i}^i({\bm x}^{t_i}))$, as {Eq.\ref{eq:posterior}}  show. 
Meanwhile, we note the meta-features that extract the feature of datasets as ${\bm m}^i=(m_1^i,...,m_F^i)$ for task $t_i$, assuming that we only consider $F$ meta-features for each tasks.

Note that in our setting, the superscripts are always used to distinguish between different tasks. Subscripts of points are used to distinguish between different points in a same task. Subscripts of data-sets, mean functions, co-variance functions and variance functions are used to show the number of input-output pairs when considering them.

In this setting, the overall goal of transfer learning for Bayesian Optimization is to to find a global minimizer (or maximizer) of an unknown function on the target task, based on the information we know from previous estimates on $K$ tasks and current observation sets $D^{t_T}$:
\begin{equation}
    \bm{{x^{t_T}}}^*=\mathop{\arg\min}_{\bm{x}\in\mathcal{X}^{T}}f^T(\bm{x}),
\end{equation}
\section{Implementation of Transfer Learning: Overview}
\label{sec_overview}

\subsection{Transfer Learning: Opportunity and Challenge}
Traditional Bayesian optimization usually considers only one task, and requires sufficient evaluations of configurations to converge to a good result. 
Given a new task, traditional Bayesian optimization re-optimize the task from scratch. 
This process may cost a lot of time and computational resources. 

\begin{figure}[!t]
\begin{center}
\centerline{\includegraphics[width=\columnwidth]{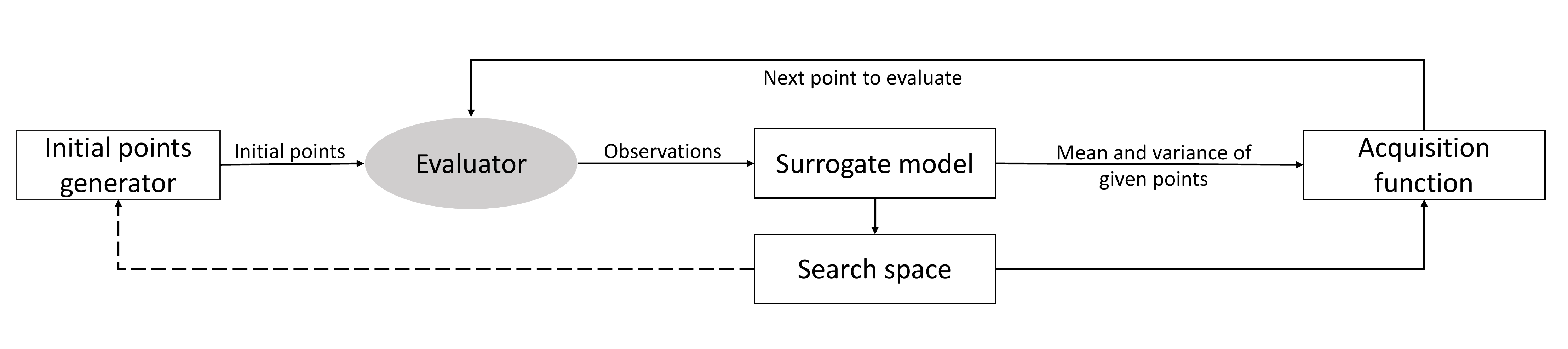}}
\caption{Bayesian Optimization Framework.}
\vspace{-2em}
\label{framework}
\end{center}
\end{figure}

In practice, researchers observed that similar tasks are likely to have similar response surface. Therefore, leveraging information from source tasks provides an opportunity to accelerate the searching process of the target task, and therefore reduce the time and computational resources.
However, leveraging information from source tasks to the target task is not simple. Challenges mainly lie in:
\begin{enumerate}
    \item How to properly use history tasks (source tasks): Before leveraging the information from history tasks to target task, it is necessary to carefully consider what to use and how to use. The first challenge is the heterogeneous scales and noise levels between different tasks. Besides, not all history tasks is helpful to the target task, it is also necessary to exclude those dissimilar tasks and only utilize those similar and helpful history tasks.
    \item How to leverage information into the target task: As {Fig. \ref{framework}} shows, there are five parts in BO framework. In order to utilize the information from history tasks, one have to choose add the information to which part in BO and make sure that this action would not lead to too much additional time and computational resource.
\end{enumerate}

\subsection{Categories of Transfer Learning-Based Bayesian Optimization}

In this survey, we propose a taxonomy to classify the existing transfer learning-based BO models for the first time. As {Fig.\ref{framework}} shows, we divide the BO framework into the following five parts:

\begin{enumerate}
    \item \textbf{Initial points generator:} The initial points generator generates initial hyper-paramater configurations for Bayesian optimization process. Previous work focuses on finding out promising initial points to accelerate the searching process, as {Sec.\ref{sec_warmstart}} introduces.
    \item \textbf{Evaluator:} The evaluator can be anything Bayesian optimization can be applied for. As we will introduce in {Sec.\ref{sec_apps}}, the evaluator can be machine learning models with certain hyperparameters, databases with various knobs to tune, etc.
    \item \textbf{Surrogate model:} As we introduced in {Sec.\ref{surrogate model}}, the surrogate model is used to give out the marginal distribution of unknown points by fitting the observations we gained before. Previous work focuses on leverage information from source tasks to help building the probabilistic model in this part. For details see {Sec.\ref{sec_surrogate}}. 
    \item \textbf{Acquisition function:} The acquisition function aims to find out next promising points to evaluate, as we introduced in {Sec.\ref{af}}. Similar to work that focuses on the Surrogate model, previous work also consider to leverage information of source tasks in the acquisition function to help searching, details are in {Sec.\ref{sec_af}}.
    \item \textbf{Search space:} The search space is an finite scope designed for the acquisition function to search for the next evaluating points. Previous work focuses on finding out a more tightened and promising search space based on the information from source tasks to help accelerating the searching process, as {Sec.\ref{sec_space}} introduces.
\end{enumerate}

\section{Transfer Learning from the view of surrogate design}
\label{sec_surrogate}
Among existing literature, most previous work focus on transfer learning techniques with specific surrogate model designs. As introduced in {Sec.\ref{surrogate model}}, those surrogate transfer methods can be categorized based on their inner surrogate design. 
While the Gaussian Process (GP) is the most common surrogate model used in BO, most of the previous surrogate transfer work depend on GP as the surrogate model~\cite{swersky2013multi,poloczek2016warm, yogatama2014efficient,joy2019flexible, shilton2017regret, ramachandran2018selecting, wang2018regret, wang2021automatic, wilson2016deep, wistuba2021few, jomaa2021transfer, iwata2021end, wistuba2021few, law2019hyperparameter, bardenet2013collaborative, salinas2020quantile, schilling2016scalable,wistuba2016two, wistuba2018scalable, feurer2018scalable, golovin2017google}. 
In addition to GP, there are also other surrogate designs, e,g, Bayesian Neural Network\cite{snoek2015scalable, perrone2018scalable, horvath2021hyperparameter, springenberg2016bayesian}, Neural Processes\cite{wei2021meta}, and Tree Parzen Estimators \cite{souza2021bayesian}.

\subsection{Gaussian Process as surrogate model}
\label{GPTL}
Gaussian Process is the most common surrogate model for Bayesian Optimization.
When considering transfer learning from the source tasks to the target task through Gaussian Processes, there exist usually three main problems to solve:
(1) how to construct the kernel function between points from different tasks;
(2) how to set the GP prior;
(3) how to deal with heterogeneous scales and noise levels between different tasks. 

The most intuitive idea for transfer learning through GP is to put the datasets from source tasks and the target task into a single GP model. 
Previous works consider kernel design, GP prior design and response surface design to solve the three main problems mentioned above respectively.
We will introduce these methods in {Sec.\ref{kernel}, \ref{prior}, \ref{response}} , respectively.

Meanwhile, some methods consider learning the individual GP models of each source task, and then learn an ensemble model based on those GP models for the target task. We will introduce these methods in {Sec.\ref{GPs}}.

\tikzset{
  FARROW/.style={arrows={-{Latex[length=1.25mm, width=1.mm]}}, }, 
  U/.style = {circle, draw=melon!400, fill=melon, minimum width=1.4em, align=center, inner sep=0, outer sep=0},
  I/.style = {circle, draw=tea_green!400, fill=tea_green, minimum width=1.4em, align=center, inner sep=0, outer sep=0},
  cate/.style = {rectangle, draw, minimum width=8em, minimum height=2em, align=left, rounded corners=3}, 
  cate2/.style = {rectangle, minimum width=2em, minimum height=2em, align=center, rounded corners=3},
  encoder/.style = {rectangle, fill=Madang!82, minimum width=10em, minimum height=3em, align=center, rounded corners=3},
}

\begin{figure}
    \centering
    \resizebox{0.9\linewidth}{!}{
    \begin{tikzpicture}
    
    \node [cate, distance=4cm, yshift=-0.5cm, align=left] (n1) at (0, 0) {\textbf{GP as surrogate model}};
    
    \node [cate, right of=n1, node distance=5cm, yshift=3cm,] (n21) {\textbf{Kernel Design}};
    \node [cate, below=3cm of n21.west, anchor=west] (n22) {\textbf{Prior Design}};
    \node [cate, below=2cm of n22.west, node distance=5cm, anchor=west,] (n23) {\textbf{Data Scale Design}};
    \node [cate, below=1.5cm of n23.west, anchor=west] (n24) {\textbf{Ensemble Design}};

    \draw[] (n1.east) -- (n21.west);
    \draw[] (n1.east) -- (n22.west);
    \draw[] (n1.east) -- (n23.west);
    \draw[] (n1.east) -- (n24.west);
    
    \node [cate, right of=n21, node distance=6cm, yshift=1cm,] (n211) {\textbf{Multi-task Kernel Design:}\\\cite{swersky2013multi},~\cite{poloczek2016warm},~\cite{yogatama2014efficient},~\cite{tighineanu2022transfer},~\cite{law2019hyperparameter}};
    \node [cate,  below=1.5cm of n211.west, anchor=west] (n212) {\textbf{Noisy Biased Kernel design:}\\Env-GP~\cite{joy2019flexible}, Diff-GP~\cite{shilton2017regret},\\
    Task Selection~\cite{ramachandran2018selecting}};
    
    \draw[] (n21.east) -- (n211.west);
    \draw[] (n21.east) -- (n212.west);

    \node [cate, below=1.5cm of n212.west, anchor=west] (n221) {\textbf{Prior GP Mean and Kernel Design:}\\
    MetaBO~\cite{wang2018regret}, HyperBO~\cite{wang2021automatic}};
    \node [cate, below=1.5cm of n221.west, anchor=west] (n222) {\textbf{Deep Kernel Prior:} \\
    DKPD~\cite{wistuba2021few}, \cite{jomaa2021transfer}, DKAF\cite{iwata2021end} 
    };
    
    \draw[] (n22.east) -- (n221.west);
    \draw[] (n22.east) -- (n222.west);

    \node [cate, below=1.5cm of n222.west, anchor=west] (n231) {\cite{yogatama2014efficient}, \cite{wistuba2021few}, \cite{bardenet2013collaborative}, \cite{salinas2020quantile}};
    {\draw[] (n23.east) -- (n231.west);}
    
    \node [cate, below=1.5cm of n231.west, anchor=west] (n241) {TST~\cite{schilling2016scalable, wistuba2016two, wistuba2018scalable}, RGPE~\cite{feurer2018scalable}, \\TransBO~\cite{li2022transbo}, Google Vizier~\cite{golovin2017google}};
    \draw[] (n24.east) -- (n241.west);

    \end{tikzpicture}}
    \caption{Summary of GP as Surrogate Model}
    \label{fig:GP-as-sur}
\end{figure}
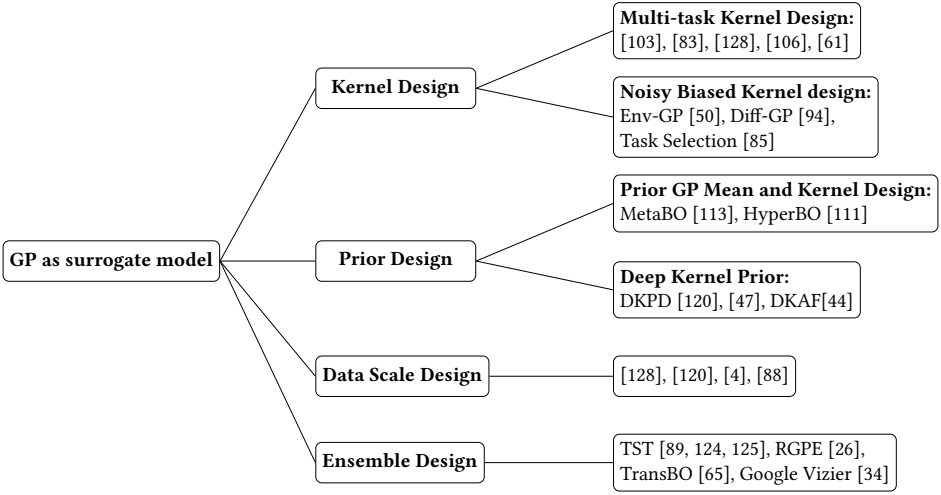

\subsubsection{Kernel Design}
\label{kernel}
Kernel function is a vital part in GP model. As {Sec.\ref{surrogate model}} introduced, for the traditional single-task BO model, the kernel function is usually pre-defined as {\em Mat\v{e}rn kernels}, {\em Squared Exponential kernel} and {\em RBF kernel}. However, in transfer learning problem, it is important to make a difference between points from source tasks and the target task. Therefore, some kernel design methods are proposed to additionally compute the difference between tasks.

Multi-task kernel design is based on the setting that considers previous observations from source tasks and target task together, and train the GP surrogate model with those observations and a designed kernel function to compute the covariance between points from different tasks.


\noindent
\textbf{A. Multi-task Kernel Design.}
Several work put source tasks and target task together into one GP model, and consider the difference between tasks to compute the kernel of this GP model. This work can be summarize as {\em Multi-task Kernel Design}.

\citet{swersky2013multi} first propose a method called {\em Multi-Task Gaussian Processes}, in which they define the multi-tasks kernel between different tasks by considering the correlation of tasks. The multi-tasks kernel in this work is called {\em intrinsic model of coregionalization} :
\begin{equation}
    K_{multi}(({\bm x}^{t_i},t_i),({\bm x}^{t_j},{t_j}))=K_t(t_i,t_j)\otimes K_x({\bm x}^{t_i},{\bm x}^{t_j}),
    \label{eq:multi-kernel}
\end{equation}
where $\otimes$ means the Kronecker product, $K_x$ means the kernel between different input points, same as the kernel in traditional GPs model, and $K_t$ measures the difference between tasks. In this work, the parameters of $K_t$ was inferred using slicing sampling, specifically, $K_t$ is represented by Cholesky factor and samples in the space. To leverage this multi-task kernel, this work assume that all tasks are positively correlated.

\citet{poloczek2016warm} also consider different tasks into single Gaussian Processes based on multi-task kernel design. They rethink about the property of covariance and deduce that for points ${\bm x}^{t_i}$ in task $t_i$ and ${\bm x}^{t_j}$ in task $t_j$, the covariance function between them can be computed as follows:
\begin{equation}
\begin{split}
    k({\bm x}^{t_i},{\bm x}^{t_j})&=Cov(f^i({\bm x}^{t_i}),f^j({\bm x}^{t_j}))\\
    &=Cov(f^t({\bm x}^{t_i})+\delta^i({\bm x}^{t_i}),f^t({\bm x}^{t_j})+\delta^j({\bm x}^{t_j}))\\
    &=Cov(f^t({\bm x}^{t_i}),f^t({\bm x}^{t_j}))+Cov(\delta^i({\bm x}^{t_i}),\delta^j({\bm x}^{t_j}))\\
    &=k^t({\bm x}^{t_{i}},{\bm x}^{t_{j}})+\mathbbm{1}_{t_{i},t_{j}}\centerdot k^{i}({\bm x}^{t_{i}},{\bm x}^{t_{j}}),
\end{split}
\end{equation}
where they assume $\bm f$ in a joint Gaussian distribution and $\delta^i({\bm x})=f^i({\bm x})-f^T({\bm x})$ is a bias with zero expectation for task $t_i$. Since they assume that $\delta^i$ and $\delta^j$ are independent iff $i\neq j$, the indicator variable $\mathbbm{1}_{t_{i'},t_{j'}}$ is one if $t_{i'}=t_{j'}$ and $t_{i'}\ne t_T$, and zero otherwise. Note that this property can be deduced only if the predetermined kernel functions are linear.

\citet{yogatama2014efficient} consider using multiple kernel to deal with points in the same task and in different tasks. They use the {\em Squared Exponential kernel} for points in the same task and a {\em Nearest Neighbor kernel} that consider points from the $n$ nearest neighbor tasks, which they find by using Euclidean distance in the dataset feature space $\mathbbm{R}^d$, and the dataset features are computed by using the previous observations of each task. Specifically, the {\em Nearest Neighbor kernel} is defined as follows:
\begin{equation}
    k({\bm x}^{t_i},{\bm x}^{t_j})=
    \begin{cases}
    1-\frac{\|{\bm x}^{t_i}-{\bm x}^{t_j}\|_2}{B}~~~~ &\text{if}~t_j \in \text{nearest neighbor task of }t_i, \\
    0  & \text{otherwise},
    \end{cases}
\end{equation}
where $B$ is a bound that $\|{\bm x}\|_2\leq B$.

\citet{tighineanu2022transfer} 
propose a method that leverage the idea of boosting in machine learning. Their method is called {\em Boosted Hierarchical GP} (BHGP), where they only consider one source task, noted as $t_s$. They add an additional term for the kernel of the query points, $k^*=k^t+\Sigma_*^{boost}$, while the additive term is computed as, $\Sigma_*^{boost}=\Sigma_{*,*}^{t_s}+\alpha_{*,t}\Sigma_{t,t}^{t_s}\alpha_{*,t}^T-\alpha_{*,t}\Sigma_{t,*}^{t_s}-\Sigma_{*,t}^{t_s}\alpha_{*,t}^T$, where $\alpha_{*,t}=k^t({\bm x}^*,{\bm X}^t)(k^t({\bm X}^t,{\bm X}^t)+\sigma_t^2\mathbbm{1})$, and $\Sigma^{t_s}$ denotes the posterior covariance matrix of the source task, which is evaluated on the points of the target task and the query points. Therefore, the kernel can be computed as follows:
\begin{equation}
    k({\bm x}^{i},{\bm x}^{j})=\sum\limits_r[{\bm W^r}]_{i,j}k^r({\bm x}^{i},{\bm x}^{j})+\delta_{{\bm x}^{i}{\bm x}^{j}}\delta_{ij}\sigma_i^2,
\end{equation}
where $i,j,r\in\{t_s,t,*\}$, $[{\bm W^r}]_{i,j}=\delta_{ir}\delta_{jr}$. Note that $\delta_{ij}=1$ if $i=j$ and $0$ otherwise for any $i$ and $j$. Through adding this boosting term, their method reduces the computational complexity comparing to the original multi-task BO.

Also aim to design a multi-task kernel function, \citet{law2019hyperparameter} consider a specific condition when utilizing BO to tune the hyperparameters of machine learning models. In this condition, each source task has two datasets, $D^{t_k}=\{(\bm{x}_j^{t_k}, y_j^{t_k})\}_{j=1}^{n^k}$ as the training data and training results of the machine learning models, and $\{(\bm{\theta}_j^{t_k}, z_j^{t_k})\}_{j=1}^{N^k}$ as the hyperparameters configurations and the noisy evaluations of the black-box function $f^k$. Note that the notations here is little different from other methods, where their goal is to find the best hyperparameters configuration for the target task, as ${\bm{\theta}^{t_T}}^*=\mathop{\arg\min}_{\bm{\theta}\in\Theta^{T}}f^T(\bm{\theta})$. 

They also assume that all the source tasks and target task follow a same class of supervised machine learning model, such that what makes the black-box function $f$ different from task to task is only relied on the structure of inputs for the machine learning model, i.e. the input datasets $D^{t_k}$. Therefore, they learn the representation of each dataset, where they decompose the dataset $D^{t_k}$ into two parts, a joint distribution of the training data, noted as $D^{t_k}\sim\mathcal{P}_{XY}^{t_k}$, and the sample size $n_k$ for task $t_k$. Specifically, they construct a feature map $\varphi(D^{t_k})$ on joint distributions for each task, where they consider three distributions for different kinds of datasets, the marginal distribution of $X$ as $P_X$, conditional distribution $P_{Y\mid X}$, and the joint distribution $P_{XY}$. They first compute the kernel mean embedding~\cite{muandet2017kernel} as follows:
\begin{equation}
\varphi(D^{t_k})=\hat{\mu}_{P_X}=\frac{1}{n_k}\sum_{l=1}^{n_k}\phi_x({\bm x}_l),
\label{px}
\end{equation}
and they compute the kernel conditional mean operator~\cite{song2013kernel},
\begin{equation}
    \hat{\mathcal{C}}_{Y\mid X}=\Phi_y^T(\Phi_x\Phi_x^T+\lambda I)^{-1}\Phi_x=\lambda^{-1}\Phi_y^T(I-\Phi_x(\lambda I+\Phi_x\Phi_x\Phi_x^T\Phi_x)^{-1}\Phi_x^T)\Phi_x,
    \label{pyx}
\end{equation}
Meanwhile, they also compute the cross covariance operator~\cite{gretton2015notes}, 
\begin{equation}
    \hat{\mathcal{C}}_{XY}=\frac{1}{n_k}\sum_{l=1}^{n_k}\phi_x({\bm x}_l)\otimes\phi_y(y_l)=\frac{1}{n_k}\Phi_x^T\Phi_y,
    \label{pxy}
\end{equation}
where $\phi_x,\phi_y$ are feature maps learned by neural networks, which is similar to the latent representation $\phi$ in {\em deep kernel learning}~\cite{wilson2016deep} (see Eq.\ref{dk}). They then use the kernel mean embedding, the kernel conditional mean operator, and the cross covariance operator to estimate the dataset $D^{t_k}$.
While $\Phi_x=[\phi_x({\bm x}_1),...,\phi_x({\bm x}_{n_k})]$, $\Phi_y=[\phi_y(y_1),...,\phi_y(y_{n_k})]$, and $\lambda$ is a regularization parameter they learned. They then flatten $\hat{\mathcal{C}}_{Y\mid X}$ and $\hat{\mathcal{C}}_{XY}$ to obtain the  feature map $\varphi(D^{t_k})$ in each condition.

Having the feature map $\varphi(D^{t_k})$ that represent the distribution of $D^{t_k}$, they then use Gaussian process or Bayesian neural network (which we will discuss in details in Sec.\ref{bnn}) to model the objective function $f$. For the GP model, they assume that $f\sim\mathcal{N}(\mu,\mathcal{C})$, and the noisy evaluation $z\mid {\bm \theta}\sim \mathcal{N}(f({\bm \theta}),\sigma^2)$. Specifically, $\mu$ is a constant, and $\mathcal{C}$ is the kernel function corresponding to $({\bm \theta},\mathcal{P}_{XY},n)$ for each task, which is computed as
\begin{equation}
    \mathcal{C}(\{{\bm \theta}_i,\mathcal{P}_{XY}^{t_i},n^{t_i}\},\{{\bm \theta}_j,\mathcal{P}_{XY}^{t_j},n^{t_j}\})=\upsilon k_{\bm \theta}({\bm \theta}_i,{\bm \theta}_j)k_p([\varphi(D^{t_i}),n_i],[\varphi(D^{t_j}),n_j]),
\end{equation}
where $\upsilon$ is a constant, and they use Mat\v{e}rn 3/2-kernel for $k_p$ and $k_{\bm \theta}$, and the parameters are optimized by using the marginal likelihood of GP.

Following multi-task ABLR (which we will introduce in Sec.\ref{bnn}), \citet{law2019hyperparameter} propose to combine their method with multi-task ABLR. The only change they made is they replace ${\bm \Phi}^k$ in Eq.\ref{multi-ablr-sur} with the vector ${\bm \gamma}=[\upsilon([{\bm \theta}_1^{t_1},\Psi^{t_1}]),...,\upsilon([{\bm \theta}_{N_1}^{t_1},\Psi^{t_1}]),...,\upsilon([{\bm \theta}_1^{t_K},\Psi^{t_K}]),..., \upsilon([{\bm \theta}_{N_K}^{t_K},\Psi^{t_K}])]$, where $\upsilon$ is a feature map, and $\Psi^{t_i}=[\varphi(D^{t_i}),n_i]$.

\noindent
\textbf{B. Noisy Biased Kernel Design}
\label{nbkd}

Also focused on the kernel construction, some works consider to view source tasks as noisy observations of target task and develop {\em Noisy biased kernel design}. \citet{joy2019flexible} first proposed a method called {\em Envelope-BO} (also known as {\em Env-GP} in related works), where they view target task in a noisy envelope of source task, and the size of the envelope depends on the correlation between source task and target task. 
This work assumes source task and target task have same covariance function $k$. 
And the covariance matrix between one source task $t_s$ and the target task is given as follows follows:
\begin{equation}
    {\bm K_*}=\begin{bmatrix}K({\bm X}^{t_s},{\bm X}^{t_s})+\sigma_s^2{\bm I}_{n_s\times n_s} & K({\bm X}^{t_s},{\bm X}^{t_T})\\K({\bm X}^{t_T},{\bm X}^{t_s}) & K({\bm X}^{t_T},{\bm X}^{t_T})+\lambda^T{\bm I}_{n_T\times n_T}\end{bmatrix},
\end{equation}
where $K({\bm X}^{t_s},{\bm X}^{t_s})$ is a $n_s\times n_s$ matrix with $K({\bm X}^{t_s},{\bm X}^{t_s})_{i,j}=k({\bm x}^{t_s}_i,{\bm x}^{t_s}_j)$ and the same goes for the rest of three matrices. 
The key is to properly design the source noise variance $\sigma_s^2$, which has to increase when the similarity between source and target task decreases. This work considers an adaptive form of $\sigma_s^2$ due to the fact that the correlation between source and target task can vary as more evaluations are observed. They place an inverse gamma distribution with parameters $\tau_0$ and $\upsilon_0$ as a prior distribution of $\sigma_s^2$, and update $\sigma_s^2$ at each iteration using the observations of the target task and the approximations of the source task on the selected points as follows:

\begin{equation}
\begin{split}
    &\sigma_s^2\sim \text{InvGamma}(\tau_0,\upsilon_0),\\
    p(\sigma_s^2\mid \{y_i^{t_T}&-\mu^s_{n_s}({\bm x}_i^{t_T}\}_{i=1}^t)\sim \text{InvGamma}(\tau_t,\upsilon_t).\\
    \label{joy}
\end{split}
\end{equation}
They use the mode of the posterior distribution as the value of source noise variance, as $\sigma_s^2=\frac{\upsilon_t}{\tau_t+1}$.


\citet{shilton2017regret} proposed a similar algorithm called {\em Diff-GP} and proved it outperform the {\em Env-GP}. The main difference between {\em Env-GP} and {\em Diff-GP} is that the former consider to update $\sigma_s^2$ to measure the correlation between tasks, while the latter consider to compute a distribution of a new function to achieve this goal.
They define a new function $g({\bm x})=f^T({\bm x})-f^s({\bm x})$ to measure the difference between source task and target task, and assume that this function follows a GP model,
\begin{equation}
\begin{split}
    &g\mid {\bm x},\bm{X^{t_T}},\Delta y({\bm X}^{t_T}) \sim \mathcal{N}(\mu_t^g({\bm x}),\lambda_t^g({\bm x})).
\end{split}
\end{equation}

The noisy observation of function $g({\bm x})$ is computed as $\Delta y({\bm x})=y^{t_T}-\mu_{n_s}^s({\bm x})$, noted as $\Delta y({\bm x})=g({\bm x})+\epsilon_g({\bm x})$, where $\epsilon_g({\bm x})\sim\mathcal{N}(0,\lambda^T+\lambda^s({\bm x}))$.
Given a data-set $\Delta y({\bm X}^{t_T})=\{\Delta y({\bm x^{t_T}_i})\}_{i=1}^{t}$, the posterior of $g$ on a given point $\bm x$ can be computed as shown in Eq.\ref{eq:posterior}, where the prior mean here is zero.

They then use the posterior mean function $\mu_t^g$ to correct the observation in source task, thus transfer the observations from source task to target task. We note the predictive mean on the target task as ${\bm \mu}^{t_T}=\{ \mu_i^{t_T}\}_{i=1}^{n_s}=\{y_i^{t_s}+\mu_t^g({\bm x}_i^{t_s})\}_{i=1}^{n_s}$. And the covariance matrix can be computed as follows:
\begin{equation}
    {\bm K}_*=\begin{bmatrix}K({\bm X}^{t_s},{\bm X}^{t_s})+{\bm \Lambda}^g & K({\bm X}^{t_s},{\bm X}^{t_T})\\K({\bm X}^{t_T},{\bm X}^{t_s}) & K({\bm X}^{t_T},{\bm X}^{t_T})+\lambda^T{\bm I}\end{bmatrix},
\end{equation}
where ${\bm \Lambda}^g$ is a diagonal matrix with ${\bm \Lambda}^g_{i,i}=\lambda^s_{n_s}({\bm x}_i^{t_s})+\lambda^g_t({\bm x}_i^{t_s})$. Therefore, they can use this covariance matrix and corrected observations to compute the posterior of GP model of the target task, as {Eq.\ref{eq:posterior}} show.

As these transfer learning methods can only transfer knowledge from one source task to the target task at each iteration, it is important to choose the right task in each iteration to ensure the efficiency. Due to this motivation, \citet{ramachandran2018selecting} propose an additional mechanism to actively select the optimal source for transfer learning based on {\em Multi-arm bandit} (MAB), and then couple it with transfer learning methods (in this work they use {\em Env-GP}\cite{joy2019flexible}). 

Specifically, they treat every source task as an arm (or a bandit), and they define a reward function that measures the benefit gained from utilizing a certain source task to the transfer learning scenario. Specifically, they assume $K$ source tasks with indexes $k=1,...K$, and they define a random variable $r_t^k$, which means the reward when the $k$-th source task is selected at iteration $t$. In this work, they set $r_t^k=-(y_t^{t_T}-\mu^k({\bm x}_t^{t_T}))^2$, where ${\bm x}_t^{t_T}$ is the point selected in iteration t and $y_t^{t_T}$ is the noisy observation on target task, $\mu^k$ is the mean function of the source task $t_k$. 

Before starting the training process of the target task, they firstly train the GPs models of all the source tasks as {Eq.\ref{eq:posterior}} show. Then they train the target task using normal BO method (in this work they use {\em Env-GP}) with mechanism of selecting the optimal source task in each iteration. Specifically, they select the optimal source task $s_t$ at $t$-th iteration by finding the solution of $s_t=\mathop{\arg\max}_{k=1,...,K}p_t^k$, where they use a weight strategy to compare the relatedness between different tasks,
\begin{equation}
    p_t^k=(1-\gamma)\frac{\omega^k(t)}{\sum_{i=1}^K\omega^i(t)}+\frac{\gamma}{K},
\end{equation}
in which $\gamma$ is a hyper-parameter chosen from $(0,1]$. And $\omega^k(t)$ is a weight variable that update in each iteration with $\omega^k(1)=1$, 
\begin{equation}
    \hat{r}^k_t=
    \begin{cases} 
    \  r^k_t/p_t^k  & \text{if}~k=s_t, \\
    \  0  & \text{otherwise},\end{cases}
\end{equation}
\begin{equation}
    \omega^k (t+1)= \omega^k (t)\exp (\frac{\gamma\cdot\hat{r}^k_t}{M}),
\end{equation}
the correlation between the $k$-th source task and the target task.
They only update the weight of selected source task in each iteration and remain the weight of other source tasks unchanged until they are selected. Finally, they prove that using this reward function, their source selection strategy based on MAB converges to the optimal source for transfer learning.

\subsubsection{Prior Design}
\label{prior}
Above methods put evaluations from source tasks and target task together to consider problem as multi-tasks or dual-tasks problem, and design the kernel that can be properly used to compute the covariance of points from different tasks. While another line consider to use the source tasks to learn the prior mean and(or) kernel function, thus implement transfer learning for BO.

\noindent
\textbf{A. GP Mean and Kernel Prior Design}

\citet{wang2018regret} propose a method called {\em MetaBO} to leverage observations from source tasks to train a prior mean and covariance function before the online training process of target task begins. They assume that all the objective functions of different tasks are sampled from the same GP prior distribution and are conditionally independent, and consider two conditions where the search space $\mathcal{X}$ is either a finite set or a compact subset of $\mathbbm{R}^d$. For both conditions, they first give out an estimator of the prior mean and kernel based on the observations from source tasks, and then give out an estimator to compute the posterior on the target task.

When the search space is a finite set $\mathcal{X}=[{\bm x}_j]_{j=1}^M$, the dataset from all source tasks can be noted as $D=\{[({\bm x}_j^{t_k},\delta_j^{t_k}y_j^{t_k})]_{j=1}^M\}_{k=1}^K$, where $y_j^{t_k}$ is computed by using the mean function from the GP model $\mathcal{N}(f^k({\bm x}),\sigma^2)$ trained on task $t_k$, in which they assume the GP model for all tasks have a same variance $\sigma$. $\delta_j^{t_k}\in\{0,1\}$ denotes whether the experiment failed to compute function value, for this value missing problem, they use the matrix completion technique in~\cite{candes2009exact} to fill in the missing value in the observation matrix ${\bm Y}=\{y_j^{t_k}\}_{j\in[M],k\in[K]}$. Then they use the unbiased estimator for the prior mean and kernel function, as $\hat{\mu}(\mathcal{X})=\frac{1}{K}{\bm Y}^T\mathbbm{1}_K$ and $\hat{k}(\mathcal{X})=\frac{1}{K-1}({\bm Y}-\mathbbm{1}_K\hat{\mu}(\mathcal{X})^T)^T({\bm Y}-\mathbbm{1}_K\hat{\mu}(\mathcal{X})^T)$, where $\hat{\mu}(\mathcal{X})\sim \mathcal{N}(\mu(\mathcal{X}),\frac{1}{K}(k(\mathcal{X})+\sigma^2))$ and $\hat{k}(\mathcal{X})\sim\mathcal{W}(\frac{1}{K-1}(k(\mathcal{X})+\sigma^2{\bm I}),K-1)$. Thus they can leverage this prior mean and kernel function trained on the source tasks to construct the GP posterior for the target task as Eq.\ref{eq:posterior}. They give out an unbiased estimator using $\hat{\mu}$ and $\hat{k}$ for the GP posterior at $t$-th iteration on the target task as follows:
\begin{equation}
\begin{split}
    \hat{\mu}_t({\bm x})&=\hat{\mu}({\bm x})+\hat{k}({\bm x },{\bm X}_t)\hat{k}({\bm X}_t,{\bm X}_t)^{-1}({\bm y}-\hat{\mu}_t({\bm X}_t)),\\
    \hat{k}_t({\bm x},{\bm x}')&=\frac{K-1}{K-t-1}\left(\hat{k}({\bm x},{\bm x}')-\hat{k}({\bm x},{\bm X}_t)\hat{k}({\bm X}_t,{\bm X}_t)^{-1}\hat{k}({\bm X}_t,{\bm x}')\right),
    \label{prior-post}
\end{split}
\end{equation}
where ${\bm X}_t=\{{\bm x}_j\}_{j=1}^t$.

When the search space is a compact subset of $\mathbbm{R}^d$, i.e. $\mathcal{X}\subset\mathbbm{R}^d$, they assume that there exists a given basis function $\Phi({\bm x})=[\phi_s({\bm x})]_{s=1}^p:\mathcal{X}\to\mathbbm{R}^p$, a mean parameter ${\bm u}\in\mathbbm{R}^p$, and a covariance parameter $\Sigma\in\mathbbm{R}^{p\times p}$, such that $\mu({\bm x})=\Phi({\bm x})^T{\bm u}$ and $k({\bm x},{\bm x}')=\Phi({\bm x})^T\Sigma\Phi({\bm x}')$. They also assume that the observations 
${\bm y}^{t_k}=\Phi({\bm X})^T{\bm W}^{t_k}+{\bm \epsilon}^{t_k}\sim\mathcal{N}(\Phi({\bm X})^T{\bm u},\Phi({\bm X})^T\Sigma\Phi({\bm X})+\sigma^2{\bm I})$, where ${\bm W}^{t_k}$ is the linear operator of task $t_k$, ${\bm W}\sim\mathcal{N}({\bm u},\Sigma)$. Then if the matrix $\Phi({\bm X})\in\mathbbm{R}^{p\times M}$ is reversible, the unbiased estimator of ${\bm W}^{t_k}$ can be given as $\hat{\bm W}^{t_k}=(\Phi({\bm X})\Phi({\bm X})^T)^{-1}\Phi({\bm X}){\bm y}^{t_k}\sim\mathcal{N}({\bm u},\Sigma+\sigma^2(\Phi({\bm X})\Phi({\bm X})^T)^{-1})$, note $\mathbb{W}=[\hat{\bm W}^{t_i}]_{i=1}^K$. Therefore, as the basis function $\Phi({\bm x})$ is given, learning the mean and kernel function $\mu$ and $k$ is equivalent to learning the mean and covariance parameter $\bm u$ and $\Sigma$. They use the estimator $\hat{\bm u}=\frac{1}{K}\mathbb{W}^T\mathbbm{1}_K$ and $\hat{\Sigma}=\frac{1}{K-1}(\mathbb{W}-\mathbbm{1}_K\hat{\bm u})^T(\mathbb{W}-\mathbbm{1}_K\hat{\bm u})$ for $\bm u$ and $\Sigma$, where $\hat{\bm u}\sim\mathcal{N}({\bm u},\frac{1}{K}(\Sigma+\sigma^2(\Phi({\bm X})\Phi({\bm X})^T)^{-1}))$ and $\hat{\Sigma}\sim\mathcal{W}(\frac{1}{K-1}(\Sigma+\sigma^2(\Phi({\bm X})\Phi({\bm X})^T)^{-1}),K-1)$. Similar to Eq.\ref{prior-post}, they give out an estimator of the posterior of the linear operator ${\bm W}\sim\mathcal{N}({\bm u}_t,\Sigma_t)$ at $t$-th iteration,
\begin{equation}
\begin{split}
    \hat{\bm u}_t&=\hat{\bm u}+\hat{\Sigma}\Phi({\bm X}_t^{t_T})(\Phi({\bm X}_t^{t_T})^T\hat{\Sigma}\Phi({\bm X}_t^{t_T}))^{-1}({\bm y}_t^{t_T}-\Phi({\bm X}_t^{t_T})^T{\bm u}),\\
    \hat{\Sigma}_t&=\frac{K-1}{K-t-1}(\hat{\Sigma}-\hat{\Sigma}\Phi({\bm X})(\Phi({\bm X}_t^{t_T})^T\hat{\Sigma}\Phi({\bm X}_t^{t_T}))^{-1}\Phi({\bm X}_t^{t_T})^T\hat{\Sigma}),
\end{split}
\end{equation}
where ${\bm X}_t^{t_T}$ denotes the queried points at $t$-th iteration on the target task. Then the posterior mean and variance on a given point $\bm x$ can be given as $\hat{\mu}_t({\bm x})=\Phi({\bm x})^T\hat{\bm u}_t$ and $\hat{k}_t({\bm x})=\Phi({\bm x})^T\hat{\Sigma}_t\Phi({\bm x})$. Moreover, for both conditions they shows that the regret bounds hold for BO.\\

Building upon the method proposed by \citet{wang2018regret}, \citet{wang2021automatic} propose {\em HyperBO}, while the former work requires that for all tasks the input points are same, which is not required for the latter work. Their method views Bayesian optimization as a parameter-led process, where they assume the task is defined by a parameter $\theta\sim p(\theta\mid\alpha)$ and the variance $\sigma\sim p(\alpha\mid \theta)$, and the GP model $\mu,k\sim p(\mu,k\mid \theta)$. Then as they make similar assumption as~\cite{wang2018regret}, the objective functions for all tasks are viewed sampled independently from the GP model with $\mu$ and $k$. Therefore, in their view, their transfer method is to leverage the previous source tasks to determine the parameters in $\mu,k,\sigma$, and set them as the GP prior for the target task. Specifically, they use two method to determine the parameters, which is either to marginal the log likelihood, or considering an empirical divergence between their defined multivariate Gaussian estimators and the true model predictions.

\noindent
\textbf{B. Deep Kernel Prior}

To learn a proper kernel function, some researchers consider the idea of {\em deep kernel learning}~\cite{wilson2016deep}. The main difference between deep kernels and the traditional kernels 
is that traditional kernels give out the form of well-defined kernels and the hyper-parameters are learnt through training process, 
while deep kernels use a neural network $\phi$ to learn a latent representation of $\bm x$, and then use this latent representation to define a kernel function as follows:
\begin{equation}
    k_{deep}({\bm x},{\bm x}'\mid {\bm \theta},{\bm \omega})=k(\phi({\bm x},{\bm \omega}),\phi({\bm x}',{\bm \omega})\mid {\bm \theta}).
    \label{dk}
\end{equation}

Based on this idea, \citet{wistuba2021few} and \citet{jomaa2021transfer} develop the {\em Deep kernel prior design}. They consider to leverage a collection of source tasks using the few-shot learning technique to learn the hyper-parameters of the deep kernel ({Eq.\ref{dk}}), thus transfer the parameterized deep kernel to the target task. Specifically, they use the estimates $\hat{\bm \theta}$ and $\hat{\bm \omega}$ to approximate the conditional distribution of $f$,
\begin{equation}
    p(f\mid{\bm x},D)=\int p(f\mid{\bm x},{\bm \theta},{\bm \omega})p({\bm \theta},{\bm \omega}\mid D)d{\bm \theta},{\bm \omega}\approx p(f\mid{\bm x},D,\hat{\bm \theta},\hat{\bm \omega}).
    \label{deepkernel}
\end{equation}
They use stochastic gradient ascent (SGA) to maximize the marginal likelihood of this distribution at each iterations. Specifically, they use a batch of observations from one sampled source task at each , to update the hyper-parameters, and get the final estimates after a given iteration time T.

\citet{iwata2021end} also develop a deep kernel by combining RBF kernel with neural network as Eq.\ref{dkaf} shows. Their proposed method is called {\em Deep Kernel Acquisition Function} (DKAF). Their model contains three components, a neural network-based kernel, a Gaussian process, and a mutual information based acquisition function. They define the RBF deep kernel as follows:
\begin{equation}
    k({\bm x},{\bm x}'\mid {\bm \theta},{\bm \omega})=\alpha\cdot exp\left(-\frac{1}{2\eta}\|\phi({\bm x},{\bm \omega})-\phi({\bm x}',{\bm \omega})\|^2\right)+\beta\cdot\delta({\bm x},{\bm x}'),
    \label{dkaf}
\end{equation}
where ${\bm \theta}=\{\alpha,\beta,\eta\}$ are parameters of the deep kernel, while $\phi(\cdot,{\bm \omega})$ is the neural network with parameter ${\bm \omega}$. Different from the method proposed above by \citet{wistuba2021few}, they learn the parameters by treating BO process as {\em Reinforcement Learning} (RL) process and train the parameters in neural networks and kernel using the source tasks. Specifically, for each iteration, they first randomly sample a source task, and they run Bayesian optimization using their designed Gaussian process with RBF deep kernel (as Eq.\ref{dkaf} shows) to get the mean and variance function. They convert the BO problem to RL setting, where evaluated data points is set as the state, point to be evaluated next is the action, and a gap between true maximum value and the maximum value at currently evaluated point is set as the negative reward. In this setting, they train the parameters by using RL algorithm, i.e. leveraging the {\em Policy Gradient} method~\cite{sutton1999policy} and updating the parameters by minimizing loss using a stochastic gradient method.

\subsubsection{Data Scale Design}
\label{response}
To deal with the problem of heterogeneous scale and noise levels between different tasks, previous works consider to reconstruct the response surface. The most intuitive idea to solve the scaling problem is simply standardize the function surface to a same scale. \citet{yogatama2014efficient} simply standardize the response surface for task $t_i$ by 
$y_j^{t_i}=\frac{f^i({\bm x}_j)-\mu^i}{\sigma^i},$ where they use estimates $\hat{\mu}^i=\frac{1}{n_i}\sum_{j=1}^{n_i}f^i({\bm x}_i)$ and $\hat{\sigma}^i=\sqrt{\frac{1}{n_i}\sum_{j=1}^{n_i}(f^i({\bm x}_i)-\hat{\mu}^i)^2}$ to approximate $\mu^i$ and $\sigma^i$. This technique is also applied in other work~\cite{schilling2016scalable}.

\citet{wistuba2021few} consider another method to standardize the observations. They firstly compute the maximum and minimum of observations for all tasks, noted as $y_{min}$ and $y_{max}$. For any given task $t_i$, they sample a lower limit $l^i$ and a upper limit $u^i$ from uniform distribution,
\begin{equation}
    l\sim\mathcal{U}(y_{min},y_{max}), u\sim\mathcal{U}(y_{min},y_{max})
    \label{few-stan}
\end{equation}
Then they use this lower limit and upper limit to standardize the observations of task $t_i$, as $\hat{y}_j^{t_i}=\frac{{y}_j^{t_i}-l^i}{u^i-l^i}$.

While another line take task differences into consideration, propose ranking-based response surface reconstruction methods. \citet{bardenet2013collaborative} propose \textit{Scot} algorithm to deal with this problem. They firstly define a partial order for points in $D^{t_i}_{n_i}=\{(\bm{x}_j^{t_i}, y_j^{t_i})\}_{j=1}^{n^i}$:
\begin{equation}
    {\bm x}_i^{t_k}\prec {\bm x}_j^{t_k} \iff y_i^{t_k}\leq y_j^{t_k}
\end{equation}

In their algorithm, they compute the partial order between one point and all the other points in the same task. Then they use the Gaussian process-based ranking algorithm proposed by \citet{chu2005preference} or $SVM^{RANK}$ proposed by \citet{joachims2002optimizing} to compute a reconstructed response surface $\hat{f}^i$ to estimate function value $f^i$. Different from evaluation $y$, for all tasks the estimates $\hat{f}$ are in a same scale, thus the heterogeneous scaling problem is solved.

\citet{salinas2020quantile} propose a reconstructing method based on semi-parametric Gaussian Copulas~\cite{wilson2010copula,anderson2017sample}, where they use Gaussian Copulas to map the observations from different tasks to comparable estimates. Specifically, they build a CDF $F(y)$ with Winsorized cut-off estimator as
\begin{equation}
    F(t)\approx
    \begin{cases}
    \delta_N &if~\Tilde{F}(t)< \delta_N\\
    \Tilde{F}(t) &if~\delta_N\leq\Tilde{F}(t)\leq1-\delta_N\\
    1-\delta_N &if~\Tilde{F}(t)>1-\delta_N
    \end{cases} 
\end{equation}
where $\Tilde{F}(t)=\frac{1}{N}\sum_{i=1}^N\mathbbm{1}_{y_i\leq t}$, N is the number of observations $y$. Intuitively, this CDF is to replace observation $y$ by rank and then normalize it within a given task. With this CDF and the standard normal CDF $\Phi^{-1}$, they can obtain a new variable by mapping the observations through a bijection $z=\phi(y)=\Phi^{-1}\circ F(y)$, and $z$ follows a normal distribution. Then they compute the conditional distribution of $z$,
\begin{equation}
    P(z\mid {\bm x})\sim \mathcal{N}(\mu_\theta({\bm x}),\sigma^2_\theta({\bm x})).
    \label{GCP}
\end{equation}
Specifically, they minimize the Gaussian negative log-likelihood using the observations from $K$ source tasks with stochastic gradient descent method. Therefore, they can standardize variable $z$ with deterministic functions $\mu_\theta$ and $\sigma^2_\theta$, as $\frac{\phi(y)-\mu_\theta({\bm x})}{\sigma_\theta({\bm x})}$.

\subsubsection{Ensemble Design}
\label{GPs}
To deal with problems mentioned at the beginning of this section, another line considers to build different tasks in separative GP models.

\citet{schilling2016scalable} propose POGPE to compute $K$ different source tasks as individual GP model and then compute the weighted product of all individual likelihoods:
\begin{equation}
    p({\bm y}\mid {\bm X},\theta)=\prod_{i=1}^K p_i^{\beta_i}({\bm y}^{t_i}\mid {\bm X}^{t_i},\theta^{t_i}),
\end{equation}
where they take $\beta_i=1/K$ for every $i$. Therefore, the mean and variance of the objective function on a given point can be computed as $\mu^*({\bm x}) = \lambda^*({\bm x})\sum_{i=1}^K\beta_i\cdot(\lambda^i)^{-1}({\bm x})\cdot\mu^{i}({\bm x})$ and $(\lambda^*)^{-1}({\bm x})=\sum_{i=1}^K\beta_i\cdot(\lambda^i)^{-1}({\bm x})$. For the target task, they consider two methods, either considering target task equal to the source tasks and take $\beta_T=1/(K+1)=\beta_i$ or giving a higher weight to target task as $\beta_T=1/2$ and source tasks as $\beta_i=1/K$.

\citet{wistuba2016two} consider a similar idea and build {\em Two-Stage Surrogate model}~(TST). At the first stage, they compute the individual GP models for both source tasks and target task similar to the above method. At the second stage, they take task correlation into consideration and compute the weight of each tasks as follows:
\begin{equation}
    \beta_k = \delta(n_k)\left(\frac{\|{\bm \chi}^k-{\bm \chi}^T\|_2}{\rho}\right),
    \label{tst-beta}
\end{equation}
where the Epanechnikov kernel $\delta(t)$ is $\frac{3}{4}(1-t^2)$ if $t\leq1$ and 0 otherwise, and $\rho>0$ is a given parameter that shows the bandwidth. For task $t_k$, ${\bm \chi}^k$ shows the feature of the task, which they use either meta-features or pairwise ranking. The pairwise ranking for both target task and source tasks is define as ${\bm \chi}^k$ with
\begin{equation}
    ({\bm \chi}^k)_{j+(i-1)t}=
    \begin{cases}
    1&\text{if}~\mu^k({\bm x}_i)>\mu^k({\bm x}_j)\\
    0&\text{otherwise}
    \end{cases}
    \label{tst-beta2}
\end{equation}
where $\mu^k$ is the mean function trained at the first stage. Therefore, the final mean function can be computed as $\mu^*({\bm x})=\frac{\sum_{i=1}^{K}\beta_i\mu^i({\bm x})+\beta_T\mu^T({\bm x})}{\sum_{i=1}^{K+1}\beta_i+\beta_T}$ and the variance function is simply defined as the variance of the target task $\lambda^T$.

In \cite{wistuba2018scalable}, they summarize their previous works product of GPs~\cite{schilling2016scalable} and TST~\cite{wistuba2016two}, and give out some variant of above functions.\\

\citet{feurer2018scalable} propose {\em Ranking-Weighted Gaussian Process Ensembles} (RGPE) also based on the idea of ensemble GPs. They first train the GP posterior for every source task, and use the posterior mean function to compute the the number of misranked pairs between one source task and the target task as follows:
\begin{equation}
    \mathcal{L}(f^k)=\sum\limits_{i=1}^{n_T}\sum\limits_{j=1}^{n_T}\mathbbm{1}(\mu^k({\bm x}_i^{t_T})<\mu^k({\bm x}_j^{t_T})\oplus y_i^{t_T}<y_j^{t_T}),
    \label{rgpe-loss}
\end{equation}
where $\oplus$ is the exclusive-or operator, and $\mathbbm{1}$ is 1 if the logical expression is true and 0 otherwise.
Then they use this loss to compute the rank for each source task. Specifically, they draw $S$ samples from the loss of all tasks, as $l^k_s\sim \mathcal{L}(f^k)$, and the weight of source task $t_k$ is computed as
\begin{equation}
    \beta_k =\frac{1}{S}\sum\limits_{s=1}^S\mathbbm{1}\left(k=\mathop{\arg\min}\limits_{k'}l_s^{k'}\right),
\end{equation}
where they use this sample technique to consider the overall uncertainty of the source and target models. Therefore, they can compute the mean and variance function at a given point ${\bm x}$ of the objective function as $\mu^*({\bm x})=\sum_{k=1}^K\beta_k\mu^k({\bm x})$ and $\lambda^*({\bm x})=\sum_{k=1}^K(\beta_k)^2\lambda^k({\bm x})$.

They further propose a weight dilution technique to remove useless source tasks. They set the weight as zero of given source task $t_k$ if the median of its loss samples $l^k_s$ is greater than $95th$ percentile of the loss samples of the target task $l^T_s$. Furthermore, in~\cite{feurer2018practical} they give out a more detailed version of RGPE and some variants of the above method.\\

\citet{li2022transbo} propose {\em TransBO}, which considers transfer learning for BO as a two-phase framework. Specifically, their two-phase framework learns the knowledge from source tasks in the first phase, and aggregates the knowledge from source tasks with the observations in the target task together in the second phase. Based on this two-phase framework, they can generate a combined transfer learning surrogate model that leverages both information from source tasks and the target task. And this model can help to choose the next promising configuration to evaluate in the next iteration. They formulate their whole learning process into a constraint optimization problem, which provides their algorithm theoretical guarantee.\\

\citet{golovin2017google} propose a multi-GPs model that is quite similar to the idea of {\em Noisy biased kernel design} introduced in Sec.\ref{nbkd}-B, where they also consider to train a distribution on the residual between the observed function value and the predicted value gained from the GP model trained at previous recursion. Specifically, at $k$-th iteration, they utilize the mean function trained at $k-1$-th iteration, noted as $\mu_{prior}$ to train the GP model of the residual with dataset $D^k_{residuals}=\{({\bm x}^{t_k}_i,y^{t_k}_i-\mu_{prior}({\bm x}^{t_k}_i)\}_{i=1}^{n_k}$, and the mean and variance function of this residual GP model are noted as $\mu_{top}$ and $\sigma_{top}$. And they use this residual GP model to correct the GP model trained at $k-1$-th iteration to gain the predicted mean and variance of a given new point, as $\mu({\bm x})=\mu_{prior}({\bm x})+\mu_{top}({\bm x})$, $\beta=\alpha\|D_k\|/(\alpha\|D_k+D_{k-1})$ and $\sigma({\bm x})=\sigma_{top}^\beta({\bm x})\sigma_{prior}^{1-\beta}({\bm x})$, where $\sigma_{prior}$ is the variance function gained at $k-1$-th iteration. Note that here $\beta$ is a weight in which $\alpha\approx1$ measures the the relative importance of the prior and top variance function.

\subsection{Bayesian Neural Network as Surrogate Model}
\label{bnn}
Due to the cubically-increasing computational complexity of GP mentioned in {Sec.\ref{surrogate model}}, some work consider to use Bayesian neural networks as an alternative to GP surrogate model, thus develop an scalable form of BO. 

\citet{snoek2015scalable} propose a scalable BO method for single task BO, which is called {\em Deep Networks for Global Optimization} (DNGO), where they use neural networks to learn adaptive basis functions for Bayesian linear regression. 
This work is based on {\em adaptive Bayesian linear regression} (ABLR)~\cite{bishop2006pattern}, given a dataset $\mathcal{D}=\{{\bm x}_i,y_i\}_{i=1}^n$, the predictive mean and variance of a given point $\mu_n({\bm x})$ can be computed as
\begin{equation}
\begin{split}
    \mu_n({\bm x})&=[\beta {\bm K}^{-1}{\bm \Phi}^T({\bm y}-\mu_0({\bm x}))]^T{\bm \phi}({\bm x})+\mu_0({\bm x}),\\
    \lambda({\bm x})&={\bm \phi}({\bm x})^T{\bm K}^{-1}{\bm \phi}({\bm x})+\frac{1}{\beta},
    \label{single-ablr}
\end{split}
\end{equation}
where ${\bm K}=\beta{\bm \phi}^T{\bm \phi}+\alpha{\bm I}$, ${\bm \phi}(\cdot)=[\phi_1(\cdot),...,\phi_D(\cdot)]^T$ is the outputs function from the last hidden layer of the neural network, and ${\bm \Phi}$ is the matrix with ${\bm \Phi}_{n,d}=\phi_d({\bm x}_n)$. The prior mean function $\mu_0({\bm x})$ is pre-designed containing our belief of the objective function.

\citet{perrone2018scalable} (also in \cite{perrone2017multiple}) propose a method for multiple tasks BO as an extension of DNGO. Specifically, they model the surrogate of the black box function using Bayesian linear regression for each task, as the following equations:
\begin{equation}
    \begin{split}
        {\bm y}^{t_k}\mid {\bm w}^{t_k},{\bm \theta},\beta_k&\sim\mathcal{N}({\bm \Phi}^k{\bm w}^{t_k},\beta_k^{-1}{\bm I}_{n_k}),\\
        {\bm w}^{t_k}\mid \alpha_k&\sim\mathcal{N}({\bm 0},\alpha_k^{-1}{\bm I}_D),
        \label{multi-ablr-sur}
    \end{split}
\end{equation}
where the linear regression weight ${\bm w}^{t_k}$ is treated as latent variable and integrated out, 
and $\alpha_k,\beta_k,{\bm \theta}$ are learned or given during the process. The matrix ${\bm \Phi}^k$ is similar to matrix in {Eq.\ref{single-ablr}} with ${\bm \Phi}^k_{n,d}=\phi_d({\bm x}_n^{t_k})$.
Then they give out a multi-task ABLR posterior (similar to {Eq.\ref{single-ablr}}) as follows:
\begin{equation}
\begin{split}
    \mu_n^k({\bm x})&=\frac{\beta_k}{\alpha_k}({\bm \phi}({\bm x}))^T{\bm K}_k^{-1}({\bm \Phi}^k)^T,\\
    \lambda^k_n({\bm x})&=\frac{1}{\alpha_k}({\bm \phi}({\bm x}))^T{\bm K}_k^{-1}{\bm \phi}({\bm x}),
    \label{multi-ablr}
\end{split}
\end{equation}
where similar to single-task ABLR, ${\bm \phi}(\cdot)=[\phi_1(\cdot),...,\phi_D(\cdot)]^T$. They learn the parameters $\bm \theta$ of the neural network ${\bm \phi}({\bm x})$ by minimizing the negative log marginal likelihood of multi-task ABLR as follows:
\begin{equation}
    \mathcal{L}({\bm \theta},\{\alpha_k,\beta_k\}_{k=1}^K)=-\sum\limits_{k=1}^K\log P({\bm y}^{t_k}\mid {\bm \theta},\alpha_k,\beta_k),
    \label{multi-ablr2}
\end{equation}
where ${\bm y}^{t_k}\mid {\bm \theta},\alpha_k,\beta_k\sim\mathcal{N}({\bm y}^{t_k}\mid{\bm 0},\beta_k^{-1}{\bm I}_{n_k}+\alpha_k^{-1}{\bm \Phi}^k({\bm \Phi}^k)^T)$. 

Based on multi-task ABLR, \citet{horvath2021hyperparameter} apply nested dropout~\cite{rippel2014learning} to ABLR, and develop {\em multi-task ABLR with Adaptive Complexity} (ABRAC). They propose a two-step procedure, where they use the source tasks combing nested dropout to learn prior parameters of ABLR in offline procedure, and then use normal multi-task ABLR method same as {Eq.\ref{multi-ablr} and \ref{multi-ablr2}} to deal with target task in online procedure. Their surrogate model is similar to {Eq.\ref{multi-ablr-sur}}, but they allow $\alpha_k$ to be different for different $w^{t_k}$, as $p({\bm w}^{t_k}\mid {\bm \alpha}_k)=\mathcal{N}({\bm 0},diag({\bm \alpha}_k^{-1}))$, which intuitively means that the linear regression weights can have different precision. Specifically, for offline process, they apply nested dropout to the ABLR as follows:
\begin{equation}
    p({\bm y}^{t_k}\mid {\bm w}^{t_k},{\bm \theta},\beta_k)=\prod\limits_{i=1}^d\mathcal{N}([({\bm \phi}^k)_{i_\downarrow}({\bm X}^{t_k})]^T{\bm w}^{t_k},\beta_k^{-1}{\bm I}_{n_k})^{\delta_{ib^k_t}}
\end{equation}
where $\delta_{ib^k_t}$ is the Kronecker delta, and $b^k_t\sim U(1,...,d)$ is the truncation at $t$-th iteration. 
$({\bm \phi}^k)_{i_\downarrow}(\cdot)$ means vector function $({\bm \phi}^k)_{i_\downarrow}(\cdot)=[\phi_1(\cdot),...,\phi_i(\cdot),0,...0]^T$. 
For online process, they also apply {\em Automatic Relevance Determination} (ARD) to adjust the level of sparsity of the Bayesian linear regression~\cite{tipping2001sparse,wipf2007new}.

Also following multi-task ABLR, \citet{law2019hyperparameter} propose to combine TNP with multi-task ABLR, which we have introduced in Sec.\ref{prior}.C.\\

\citet{springenberg2016bayesian} propose a method named {\em Bayesian Optimization with Hamiltonian Monte Carlo Artificial Neural Networks} (BOHAMIANN), which is based on stochastic {\em Markov
chain Monte Carlo} (MCMC) method from~\cite{chen2014stochastic}. Specifically, they define the Bayesian neural networks of multi-task model as $\hat{f}^k({\bm x};\theta_\mu)=h\left([{\bm x};\psi_k]^T,\theta_h\right)$ for task $t_k$, where $h(\cdot)$ is the output of the neural network parameterized by $\theta_h$, and $\psi_k$ is the $k$-th row of an embedding matrix $\psi\in\mathbbm{R}^{K\times L}$. 
The overall mean prior parameters of this neural networks can be noted as $\theta_\mu=[\theta_h,vec(\psi)]$, where $vec()$ means vectorization, and the variance prior parameters can be noted as $\theta_{\lambda}^i$. 
In order to compute the predictive posterior $p(f^k({\bm x})\mid {\bm x},D)$, which is hard to evaluate with the choice of using neural networks, they propose to use {\em Stochastic Gradient Hamiltonian Monte Carlo} (SGHMC) to sample $\theta^i\sim p(\theta\mid D)$~\cite{chen2014stochastic}, and approximate the posterior with a summation as
    $p(f^k({\bm x})\mid {\bm x},D)=\int_\theta p( f^k({\bm x})\mid {\bm x},\theta)p(\theta\mid D)d\theta\approx\frac{1}{M}\sum_{i=1}^Mp( f^k({\bm x})\mid {\bm x},\theta^i)$.
The posterior mean and variance can be given as
\begin{equation}
\begin{split}
    \mu_n^k({\bm x})&=\frac{1}{M}\sum_{i=1}^M\hat{f}^k({\bm x};\theta_\mu^i),\\
    \lambda^k_n({\bm x})&=\frac{1}{M}\sum_{i=1}^M\left(\hat{f}^k({\bm x};\theta_\mu^i)-\mu_n^k({\bm x}))\right)^2+\theta_{\lambda}^i.
\end{split}
\end{equation}

\subsection{Neural Process as surrogate models}
\label{np}
Recent works also consider to leverage {\em Neural Processes} (NPs)~\cite{garnelo2018conditional,garnelo2018neural,kim2019attentive} as a replacement of GPs model to deal with the inefficiency problem of GPs~\cite{wei2021meta}. NPs combine the best of both neural networks and GPs to simultaneously trained with backpropagation and in a distribution, and have been proved successful in recent studies~\cite{garnelo2018conditional,wei2021meta}. 

\citet{wei2021meta} propose to use NPs as surrogate models and develop the transfer learning scenario. Their method is called {\em Transfer Neural Processes} (TNP). The Neural Process usually contains three components, an encoder $\mathcal{E}_{{\bm \theta}_e}$ that learns an embedding for every observation, a data-aware attention unit $\mathcal{A}_{{\bm \theta}_a}$ that considers all the previous observations and gives a representation of them that has invariant order, and a decoder $\mathcal{D}_{{\bm \theta}_d}$ that compute the predicted mean and variance for a given point. The encoder and the decoder are both parameterized by neural networks. 

Formally, the neural process model can be noted as $TNP_{\bm \theta}=\mathcal{E}_{{\bm \theta}_e}\circ \mathcal{A}_{{\bm \theta}_a}\circ \mathcal{D}_{{\bm \theta}_d}$. 
The overall parameters are noted as ${\bm \theta}={\bm \theta}_e\cup{\bm \theta}_a\cup{\bm \theta}_d$. Take the target task as an example, at $t$-th iteration the target task contains $n_T+t$ points that the first $n_T$ points are initialized points (we will discuss the initialization technique later) and the last $t$ points are points queried at each iteration. Following the idea from~\cite{finn2017model}, this work randomly shuffles the observations into two parts, noted as $D_{t,h}^{t_T}=\{({\bm x}_i^{t_T},y_i^{t_T})\}_{i=1}^{h}$ and $D_{t,\bar{h}}^{t_T}=\{({\bm x}_i^{t_T},y_i^{t_T})\}_{i=h+1}^{n_T+t}$, and then the conditional log likelihood is
\begin{equation}
    \mathcal{L}(D_{t,h}^{t_T}\mid D_{t,\bar{h}}^{t_T},{\bm \theta})=\mathbbm{E}_{f\sim P}[\mathbbm{E}_{h}[\log p_{\bm \theta}(\{y_i^{t_T}\}_{i=1}^h\mid D_{t,\bar{h}}^{t_T},\{{\bm x}_i^{t_T}\}_{i=1}^h)]],
    \label{np-loss}
\end{equation}
where the gradient of $\mathcal{L}$ is empirical estimated by sampling $f$ and different values of $h$. 

The specific forms of three components are given as following. Note the learned embedding of the encoder at point ${\bm x}_i^{t_k}$ as ${\bm r}_i^{t_k}=\mathcal{E}_{{\bm \theta}_e}({\bm x}_i^{t_k},y_i^{t_k})$, and the data-aware attention unit is computed in this work for a given point ${\bm x}$ as
\begin{equation}
\begin{split}
    {\bm r}_*({\bm x})&= \mathcal{A}_{\theta_a}({\bm R}^{t_T},\{{\bm r}_i^{t_1}\}_{i=1}^{n_1},...,\{{\bm r}_i^{t_K}\}_{i=1}^{n_K};{\bm x})\\
    &={\textbf{MultiHead}}(g({\bm x}),g({\bm X}),{\bm R},{\bm s})\\
    &=[{\textbf{head}}_d(g({\bm x}),g({\bm X}),{\bm R},{\bm s})]_{d=1}^{\lfloor r/d\rfloor},
    \label{attention}
\end{split}
\end{equation}
where ${\bm R}=[{\bm R}^{t_T};\{{\bm r}_i^{t_1}\}_{i=1}^{n_1},...,\{{\bm r}_i^{t_K}\}_{i=1}^{n_K}]$, and ${\bm X}=[{\bm X}^{t_T}; {\bm X}^{t_1},...,$ ${\bm X}^{t_K}]$. As previously discussed in Eq.\ref{np-loss}, when training the parameters $\theta$ by conditioning on $D_{t,h}^{t_T}$, the datasets are ${\bm R}^{t_T}=\{{\bm r}_i^{t_T}\}_{i=h+1}^{n_T+t}$, ${\bm X}^{t_T}=\{{\bm x}_i^{t_T}\}_{i=h+1}^{n_T+t}$ and ${\bm x}\in D_{t,h}^{t_T}$, while when making prediction for a given target points ${\bm x}$, the datasets are ${\bm R}^{t_T}=\{{\bm r}_i^{t_T}\}_{i=1}^{n_T+t}$ and ${\bm X}^{t_T}=\{{\bm x}_i^{t_T}\}_{i=1}^{n_T+t}$. 

In Eq.\ref{attention}, each head of the multi-head attention is computed as 
\begin{equation}
\textbf{head}_d={\rm softmax}({\bm s}\circ[g({\bm x})]^T{\bm W}_d^1[{\bm W}_d^2]^Tg({\bm X})/\sqrt{r}),
\end{equation}
where ${\bm W}_d^1,{\bm W}_d^2\in\mathbbm{R}^{r\times d}$ are parameters. Different from previous works that considers to measure the similarity between source tasks and target task through meta-features~\cite{feurer2015initializing,kim2017learning} or partial relationship~\cite{bardenet2013collaborative}, this work computes similarity through cosine similarity. Specifically, similarity ${\bm s}={\rm softmax}([\mathbbm{1}_{1\times(n_T+t-h)},s^{t_1}\mathbbm{1}_{1\times n_1},...,s^{t_K}\mathbbm{1}_{1\times n_K}])$, where $\mathbbm{1}_{1\times n_k}$ is the one vector with $n_k$ elements, and the similarity between the target task and the task $t_k$ is computed through cosine similarity as $s^{t_k}=\frac{1}{n_T+t}\sum_{i=1}^{n_T+t}{\rm cos}({\bm r}_i^{t_T},\frac{1}{Q}\sum_j^Q{\bm r}_j^{t_k})$. 
Thus given the representation $r_*$ and a given point $\hat{{\bm x}}_j$, the decoder output the predicted mean and variance of the function value $y_j$ at point ${\bm x}$, as $\mu({\bm x}),\sigma({\bm x})=\mathcal{D}_{{\bm \theta}_d}({\bm r}_*,{\bm x})$.

Before training on the target task, in order to achieve transfer from source tasks to target task, the parameters ${\bm \theta}$ are updated through pre-training on $K$ source tasks. Specifically, the parameters ${\bm \theta}$ are randomly initialized as $\Tilde{{\bm \theta}}$ and updated at $k$-th iteration (which means that the source task $t_k$ is taken into consideration, and the following text has same meaning) as
\begin{equation}
    {\bm \theta}_p^k=\Tilde{{\bm \theta}}-\alpha\bigtriangledown_{\bm \theta}^p\mathcal{L}(D_{n_k,h}^{t_k}\mid D_{n_k,\bar{h}}^{t_k},\theta),~
    \Tilde{{\bm \theta}}=\Tilde{{\bm \theta}} +\epsilon({\bm \theta}_p^k-\Tilde{{\bm \theta}})
\end{equation}
where $k=1,...,K$, $p$ denotes $p$ gradient steps, and the loss $\mathcal{L}$ is similar to {Eq.\ref{np-loss}}. At $t$-th iteration when training on the target task, they also update the parameters as ${\bm \theta}_p=\Tilde{{\bm \theta}}-\alpha\bigtriangledown_{\bm \theta}^p\mathcal{L}(D_{t,h}^{t_T}\mid D_{t,\bar{h}}^{t_T},{\bm \theta})$.\\

This work also considers a warm start method, their work has a similar idea with~\cite{wistuba2015learning} in Sec.\ref{sec_warmstart}, which also considers to use source tasks to update a set of randomly initial points. Specifically, a set of randomly initialized points $\{\Tilde{{\bm x}}_i\}_{i=1}^{n_T}$ are updated through pre-training on $K$ source tasks, at $k$-th iteration, the points are update as
\begin{equation}
    {\bm x}_i^p=\Tilde{{\bm x}}_i-\alpha\bigtriangledown_{{\bm x}_i}^p\mathcal{L}(\{\Tilde{{\bm x}}_i\}_{i=1}^{n_T}\mid\theta),~
    \Tilde{{\bm x}}_i=\Tilde{{\bm x}}_i+\epsilon({\bm x}_i^p-\Tilde{{\bm x}}_i)
    \label{np-warmstart}
\end{equation}

In this equation, the loss can be noted as 
\begin{equation}
    \mathcal{L}(\{\Tilde{{\bm x}}_i\}_{i=1}^{n_T}\mid\theta)=\sum_{i=1}^{n_T}\frac{\exp(\alpha\cdot\mu^{k}({\Tilde{{\bm x}}_i}))}{\sum_{j=1}^{n_T}\exp(\alpha\cdot\mu^{k}({\Tilde{{\bm x}}_j}))}\mu^{k}({\Tilde{{\bm x}}_i})+\frac{\exp(\alpha\cdot\sigma^{k}({\Tilde{{\bm x}}_i}))}{\sum_{j=1}^{n_T}\exp(\alpha\cdot\sigma^{k}({\Tilde{{\bm x}}_j}))}\sigma^{k}({\Tilde{{\bm x}}_i}),
\end{equation}
where along with the update of parameters $\theta$, the mean and variance update at each iteration, which we note the mean and variance function at $k$-th iteration as $\mu^k,\sigma^k$.
This loss ensures that at least one initial point has the maximum mean and variance.

\section{Transfer Learning from the view of Acquisition function design}
\label{sec_af}
While the above methods consider to implement transfer learning techniques through surrogate model, some works also consider to transfer knowledge through acquisition function. This idea arises as the transfer surrogate methods sometimes hard to deal with scaling problems. Meanwhile, surrogate transfer methods usually neglect the fact that as more new observations are gained in target task, the knowledge from source tasks become less important. Transferring knowledge through acquisition function can avoid those two problems. Previous work considers transfer learning for acquisition function from the view of multi-task BO~\cite{swersky2013multi,moss2020mumbo}, ensemble-GPs~\cite{wistuba2018scalable}, and reinforcement learning~\cite{volpp2019meta}.

\subsection{Multi-task BO acquisition function}
\citet{swersky2013multi} consider an acquisition function for multi-task BO based on entropy search. 
Specifically, they take cost into consideration, and encourage evaluating configurations with large information gained on the target task and low evaluation costs on the current task. The {\em information gain per unit cost} is as a variant of Eq.\ref{es-al}, which is computed as follows:
\begin{equation}
    \alpha_{ES_n}({\bm x}^{t_k})=\int\int\left(\frac{H({\bm x}^*\mid D_n)-H({\bm x}^*\mid D_{n+1})}{c_k({\bm x})}\right)p(y\mid f)p(f\mid {\bm x})dydf,
    \label{multi-af}
\end{equation}
where $H(\cdot)$ is the entropy on the target task, and we note $D_{n+1}=D_n\cup \{({\bm x},\mu_n({\bm x}))\}$. As multi-task BO method considers all tasks into one GP model, $D_0$ contains all data from source tasks. And $c_k({\bm x})$ is the real valued cost function of evaluating ${\bm x}$ at task $t_k$.

\citet{moss2020mumbo} also consider a variant of entropy search for multi-task BO, they call it {\em MUlti-task Max-value Bayesian Optimization} (MUMBO). Their work is based on {\em Max-value Entropy Search} (MES)~\cite{wang2017max}, and they also take fidelity ${\bm z}$ into consideration. The general form of MUMBO is
\begin{equation}
    \alpha_{MUMBO_n}({\bm x}^k,{\bm z})=\frac{H(y\mid D_n)-\mathbb{E}_{f({\bm x})}[H(y\mid D_n\cup \{({\bm x},\mu_n({\bm x}))\})]}{c_k({\bm x},{\bm z})},
    \label{mumbo}
\end{equation}
which is similar to {Eq.\ref{multi-af}}, $c_k({\bm x},{\bm z})$ is the real valued cost function of evaluating ${\bm x}$ with fidelity ${\bm z}$ at task $t_k$. Besides, based on MES, they give out a computational form of {Eq.\ref{mumbo}}, as
\begin{equation}
\begin{aligned}
    \alpha_{MUMBO_n}({\bm x}^k,{\bm z})=\frac{1}{c_k({\bm x},{\bm z})}\cdot\frac{1}{N}\sum\limits_{f^*\in G}\rho({\bm x},{\bm z})^2\frac{\gamma_{f^*}({\bm x})\phi(Z_{f^*}({\bm x}))}{2\Phi(\gamma_{f^*}({\bm x}))}
    -\\\log(\Phi(\gamma_{f^*}({\bm x})))+\mathbb{E}_{\theta\sim Z_{f^*}({\bm x},{\bm z})}\left[\log\left(\Phi\left\{\frac{\gamma_{f^*}({\bm x})-\rho({\bm x},{\bm z})\theta}{\sqrt{1-\rho({\bm x},{\bm z})^2}}\right\}\right)\right],
\end{aligned}
\end{equation}
where $\Phi$ is the standard normal cumulative distribution and $\phi$ is the probability density functions, $\gamma_{f^*}({\bm x})=\frac{f^*-\mu_f({\bm x})}{\sigma_f({\bm x})}$ and $f^*=\mathop{\max}_{{\bm x}\in \mathcal{X}}f({\bm x})$, $Z_{f^*}({\bm x},{\bm z})$ is an extended-skew Gaussian (ESG) (for details see~\cite{moss2020mumbo}).

\subsection{Ensemble GPs-based acquisition function transfer}
\citet{wistuba2018scalable} propose a method similar to TST~\cite{wistuba2016two}, but consider to transfer knowledge within acquisition function instead of surrogate model, which is called {\em Transfer Acquisition Function} (TAF). In this method, they trian the individual GP models for each source task same as TST, then they consider to leverage these knowledge to measure the improvement of a new point ${\bm x}$ through a variant of EI acquisition function (introduced in {Sec.\ref{af}}):
\begin{equation}
    \alpha_{TAF}({\bm x})=\frac{\beta_{T}\alpha_{EI_n}^T({\bm x})+\sum_{i=1}^K\beta_iI_i({\bm x})}{\sum_{i=1}^K\beta_i},
\end{equation}
where $\beta_k$ is same as how TST sets, see {Eq.\ref{tst-beta}} and {Eq.\ref{tst-beta2}}, and $I_k({\bm x})=max\{y_{min}^{t_k}-\mu^k({\bm x}),0\}$. The acquisition function $\alpha_{EI_n}^T({\bm x})$ is the EI with observations from target task at $n$-th iteration.\\

\subsection{Reinforcement learning-based acquisition function transfer}

\citet{volpp2019meta} consider the condition that the objective function of the target task share similar structure with the objective functions of the source tasks, while the source tasks are much cheaper to evaluate. They propose a hand-designed acquisition function called {\em Neural Acquisition Function} (NAF) to achieve meta learning from the source tasks to the target task. Concretely, NAF is parameterized by a vector $\theta$, noted as $\alpha_{t,\theta}$. They use the {\em Proximal Policy Optimization} (PPO) algorithm from {\em Reinforcement Learning} (RL) to learn the vector in NAF. 
\citet{hsieh2021reinforced} also propose a method using Reinforcement Learning. Different from \citet{volpp2019meta}, their method relies on deep Q-network (DQN) as differentiable surrogate of AF.

\section{Transfer Learning from the view of initialization design}
\label{sec_warmstart}
As the efficiency of BO depends on the initial points of the searching process, some works consider to find proper initial points based on previous knowledge. These works can be summarized as warm-start methods. To find the proper initial points, previous works focuses on three main direction, measuring datasets similarities and choose initial points based on meta-features~\cite{feurer2015initializing,feurer2015efficient,kim2017learning}, generating initial points by gradient-based learning~\cite{wistuba2015learning,wei2021meta}, or generating initial points using evolutionary algorithm\cite{wistuba2021few}.

\subsection{Meta-features-based initialization}
\citet{feurer2015initializing} first propose an initialization method for BO which is called {\em Meta-learning-based Initialization Sequential Model-based Bayesian Optimization} (MI-SMBO), which can be a plug-in component for other different BO methods. They apply their initialization technique to the state-of-the-art SMBO method at that time, Spearmint and SMAC, using a comprehensive suite of 57 classification datasets and 46 meta-features, and gain significant improvements. Their method considers a offline training to compute the meta-features of different source tasks (one known dataset is viewed as one source task), and the best point of each source, noted as $\hat{{\bm x}}^{t_1},...,\hat{{\bm x}}^{t_K}$ for datasets $D^{t_1},...,D^{t_K}$. Before training on the target task (new dataset), they first compute the meta-features of the target task, then compute the distance between each source task and the target task to measure the similarity between them, using meta-features with p-norm distance 
\begin{equation}
    d_p(D^{t_k},D^{t_j})=\|{\bm m}^{k}-{\bm m}^j\|_p,
    \label{p-norm}
\end{equation}
or negative Spearman correlation coefficient (Eq.\ref{spearman})
\begin{equation}
    d_c(D^{t_k},D^{t_j})=1-\text{Corr}([f^k({\bm x}_1),...,f^k({\bm x}_n)],[f^j({\bm x}_1),...,f^j({\bm x}_n)]),
    \label{spearman}
\end{equation}
Finally, they sort the distance from small to large, and select the top $t$ best points from the first $t$ datasets as sorted, i.e. they choose the best points from $t$-nearest source tasks as the initial points for the target task.

\citet{feurer2015efficient} add a warm-start component to their Automated Machine Learning (AutoML) system, where the warm-start method is quite similar to the method they proposed above~\cite{feurer2015initializing}, which is also based on meta-features but considers its application in specific condition.

Also relied on meta-features to measure datasets similarity, especially similarity between image datasets, and determine $t$-nearest source tasks, \citet{kim2017learning} propose to learn meta-features over datasets using their trained deep feature and meta-feature extractors. They first randomly sample $\tau$ data from each dataset as a subsets to reduce computational complexity. As their work considers datasets that are all image datasets, their proposed framework first extracts features of those image data by using a deep feature extractor $\mathcal{M}^{df}$, which is a deep neural networks, and output deep features ${\bm d}^{t_k}_{1:\tau}=\{{\bm d}^{t_k}_i\}_{i=1}^{\tau}$ for task $t_k$. Then the deep features are fed into a meta-feature extractor $\mathcal{M}^{mf}$, which is either {\em Aggregation of Deep Features} (ADF) 
\begin{equation}
    {\bm h}^{t_k}:={\bm h}_{ADF}^{t_k}=\sum_{i=1}^{\tau}{\bm d}_i^{t_k}~~or~~\frac{1}{\tau}\sum_{i=1}^{\tau}{\bm d}_i^{t_k},
    \label{adf}
\end{equation}
or {\em Bi-directional Long Short-Term Memory network} (Bi-LSTM) 
\begin{equation}
    {\bm h}^{t_k}:={\bm h}^{t_k}_{Bi-LSTM}={\rm Bi-LSTM}({\bm d}^{t_k}_{1:\tau}),
    \label{bi-lstm}
\end{equation}
and the output is ${\bm h}^{t_k}$ for task $t_k$.

Finally, there exists a fully-connected layer after the meta-feature extractor to produce a meta-feature vector for each tasks as ${\bm m}^k$ for task $t_k$. The parameters in their models are trained by minimizing $\|d_{target}(D_i,D_j)-d_{mf}({\bm m}^i, {\bm m}^j)\|$, where $d_{target}(D_i,D_j)=\sum_{s=1}^n\|f^i({\bm x}_s)-f^j({\bm x}_s)\|$, which shows the difference between two datasets. It is obvious that this method assumes that the response surfaces of objective function for all tasks are in a same scale.

\subsection{Gradient-based learning initialization}

\citet{wistuba2015learning} (also in \cite{wistuba2015learning2}) propose a method that does not depend on meta-feature, but can directly learn the optimal initial points through iteration. They learn a set of initial points by minimizing a defined meta loss,
\begin{equation}
    \mathcal{L}({\bm X}^I,\mathcal{D})=\frac{1}{K}\sum_{D^{t_k}\in\mathcal{D}}\mathop{\min}_{{\bm x}\in {\bm X}^I}f^k({\bm x}),
    \label{metaloss1}
\end{equation}
where ${\bm X}^I=\{{\bm x}_1,...,{\bm x}_I\}$ denotes the set of initial points that contains $I$ points, dataset $\mathcal{D}=\{D^{t_1},...,D^{t_K}\}$ is the dataset that contains all $K$ datasets of the source tasks. This meta loss is not differentiable, thus this work propose to use differentiable softmin function to approximate it,
\begin{equation}
    \mathcal{L}({\bm X}^I,\mathcal{D})=\frac{1}{K}\sum_{D^{t_k}\in\mathcal{D}}\sum_{i=1}^I\sigma_{D^{t_k},i}\hat{f}^k({\bm x}_i),
    \label{metaloss2}
\end{equation}
where $\hat{f}^k=\mu^k$ is the mean function from the GP model of task $t_k$, $\sigma_{D^{t_k},i}=\frac{\exp(\beta\hat{f}^k({\bm x}_i))}{\sum_{j=1}^I\exp(\beta\hat{f}^k({\bm x}_j))}$, in which they choose $\beta=-100$ such that the summation $\sum_{i=1}^I\sigma_{D^{t_k},i}\hat{f}^k({\bm x}_i)$ is close to ${\rm min}\{\hat{f}^k({\bm x}_1),...,\hat{f}^k({\bm x}_I)\}$. In this form, the meta loss is differentiable. And the initial points can be randomly initialized and updated as $x_{i,j}=x_{i,j}-\eta\frac{\partial}{\partial x_{i,j}}\mathcal{L}({\bm X}^I,\mathcal{D})$, where $x_{i,j}$ is the $j$-th element of the vector ${\bm x}_i$. Moreover, they also propose an adaptive form to take dataset similarity into consideration, as the following equation:
\begin{equation}
    \mathcal{L}({\bm X}^I,\mathcal{D})=\frac{1}{K}\sum_{D^{t_k}\in\mathcal{D}}c(D^{t_k},D^{t_T})\sum_{i=1}^I\sigma_{D^{t_k},i}\hat{f}^k({\bm x}_i),
    \label{metaloss3}
\end{equation}
where $c(D^{t_k},D^{t_T}):=\frac{\sum\limits_{{\bm x}_i,{\bm x}_j\in{\bm X}^I}s({\bm x}_i,{\bm x}_j,D^{t_k},D^{t_T})}{\|{\bm X}^I_t\|(\|{\bm X}^I_t\|-1)}$, and the similarity $s$ is defined by the partial relationship $s({\bm x}_i,{\bm x}_j,D^{t_k},D^{t_T}):=\mathbbm{1}(\hat{f}^k({\bm x}_i)>\hat{f}^k({\bm x}_j)\oplus f^T({\bm x}_i)>f^T({\bm x}_j))$ similar to Eq.\ref{rgpe-loss}, which shows the number of misranked pairs.

As Sec.\ref{np} has introduced, \citet{wei2021meta} also propose a warm start method based on Neural Processes model, which is similar to the method above proposed by \citet{wistuba2015learning}, see Eq.\ref{np-warmstart} for more details.

\subsection{Evolutionary algorithm based initialization}
\citet{wistuba2021few} propose a warm start method based on evolutionary algorithm. They use an evolutionary algorithm to find a set of points that can minimize the loss on the source tasks, as the following equation:
\begin{equation}
    {\bm X}^I=\mathop{\arg\min}_{{\bm X}\subseteq\mathcal{X}}\sum_{k=1}^K\mathcal{L}(f^k,{\bm X})=\mathop{\arg\min}_{{\bm X}\subseteq\mathcal{X}}\sum_{k=1}^K \underset{{\bm x}\in{\bm X}}{\min}\frac{f^k({\bm x})-f^k_{\min}}{f^k_{\max}-f^k_{\min}},
\end{equation}
where $f^k_{min}$ and $f^k_{max}$ are the minimum and maximum of the function values considering all points estimated so far (in $\bm X$ and $D^{t_k}$), while the function value at a previously unobserved point for the task is estimated by using the mean function from the GP surrogate model trained before for each source tasks. Specifically, the evolutionary algorithm works as follow. They first sample a set of $I$ random points with sampled proportion for each point (take point $\bm x$ as an example) as follows:
\begin{equation}
    \exp\left(-\underset{k\in\{1,...,K\}}{\min}\frac{f^k({\bm x})-f^k_{\min}}{f^k_{\max}-f^k_{\min}}\right).
\end{equation}
Then the traditional evolutionary algorithm works, which randomly chooses to either do mutation for the set to replace elements with new points, or perform a crossover operation between two sets to generate new set with elements from both sets. Thus a new set is generated and added to the population. They repeat this process for $100,000$ steps to find the best set as an initial set for the target task.

\section{Transfer Learning from the view of space design}
\label{sec_space}


Apart from surrogate model, acquisition function and warm-starting, some works also consider to design a promising space for the target task based on the knowledge from source tasks\cite{wistuba2015hyperparameter,perrone2019learning,li2022transfer}. 

\subsection{Search space pruning}
\citet{wistuba2015hyperparameter} first consider a search space pruning technique, i.e. pruning unpromising space using the knowledge from source tasks to avoid unnecessary function evaluations. They define a region $\mathcal{R}$ by a center point $\bm x$ and a diameter $\delta$. They first evaluate task similarity by computing the Kendall tau rank correlation coefficient~\cite{kendall1938new},
\begin{equation}
    \text{KTRC}(D^{t_k},D^{t_T}):=\frac{\sum\limits_{{\bm x}_i,{\bm x}_j\in{\bm X}_t}\mathbbm{1}(\hat{f}^k({\bm x}_i)>\hat{f}^k({\bm x}_j)\oplus f^T({\bm x}_i)>f^T({\bm x}_j))}{\|{\bm X}_t\|(\|{\bm X}_t\|-1)},
    \label{KTRC}
\end{equation}
where the numerator shows the number of misranked pairs, and ${\bm X}_t$ is the set of already evaluated hyperparameter configurations on the target task after $t$ trials. $\hat{f}^k$ is approximated by the mean function from the GP model of task $t_k$, which is normalized to deal with the scaling problem. They select $m$ source task that are most similar to the target task and note them as $T'=\{t_{i1},....,t_{im}\}$. Then they compute the defined potential that shows how promising a search space is, as the following equation:
\begin{equation}
    \text{potential}(\mathcal{R}=({\bm x},\delta),{\bm X}_t):=\sum_{k \in \{i1,...,im\}}\hat{f}^k({\bm x})-\max\limits_{{\bm x}' \in {\bm X}_t}\hat{f}^k({\bm x}'),
\end{equation}

Based on this defined potential, they select several hyperparameters with little potential and note the set contains them as ${\bm X}'$. Note the original search space as $\bm X$, then the pruned search space is defined as:
\begin{equation}
    {\bm X}^{\text{(pruned)}}:=\{{\bm x}\in{\bm X}\mid dist({\bm x},{\bm x}')>\delta,{\bm x}' \in {\bm X}'\}.
\end{equation}

Considering the promising points, the returned space is ${\bm X}^{(pruned)}\cup\{{\bm x}\in{\bm X}\mid dist({\bm x},{\bm x}')\leqslant\delta,{\bm x}' \in {\bm X}'\}$, where the distance is defined as follows:
\begin{equation}
    \text{dist}({\bm x},{\bm x}'):=
    \begin{cases}
    \infty & \text{if $\bm x$ and $\bm x'$ differ in a categorical variable},\\
    \|{\bm x}-{\bm x}'\|& \text{otherwise},
    \end{cases}
\end{equation}
where they consider especially the condition when changing a categorical variable, which makes the loss not smoothly changed.

\subsection{Promising search space design}
While \citet{perrone2019learning} consider to design a promising search space for the target task instead of pruning the original search space. They transfer the search space estimation problem to a constraint optimization problem, as the following equation:
\begin{equation}
    \min\limits_{\theta\in\mathbbm{R}^q}\mathcal{Q}({\bm \theta})\ \text{such that for}\ k\in\{1,...,K\},{\bm x}^{t_k*}\in\hat{\mathcal{X}}({\bm \theta}),
    \label{optimization}
\end{equation}
where $\bm{{x^{t_k}}}^*=\mathop{\arg\min}_{\bm{x}\in\mathcal{X}^{t_k}}f^k(\bm{x})$, and $\hat{\mathcal{X}}\subset \mathcal{X}$ is a subset of the original search space defined by a parameter vector ${\bm \theta}$. $\mathcal{Q}({\bm \theta})$ is volume measure of the search space $\hat{\mathcal{X}}({\bm \theta})$.

Specifically, they define two shape of the search space, box or ellipsoid. For the box space, the parameter vector is ${\bm \theta}=({\bm l},{\bm u})$, and the search space is designed as $\hat{\mathcal{X}}({\bm \theta})=\{{\bm x}\in\mathbbm{R}^p \mid {\bm l}\leqslant{\bm x}\leqslant{\bm u}\}$. The constraint optimization problem in Eq.\ref{optimization} can be varied as:
\begin{equation}
    \min\limits_{{\bm l}\in\mathbbm{R}^p,{\bm u}\in\mathbbm{R}^p}\frac{1}{2}\|{\bm u}-{\bm l}\|\ \text{such that for}\ k\in\{1,...,K\},{\bm l}\leqslant{\bm x}^{t_k*}\leqslant{\bm u},
    \label{box}
\end{equation}
where ${\bm l}$ is the lowest bound of the box, and ${\bm u}$ is the highest bound of the box. This optimization has a simple form of solution, ${\bm l}^*=\mathop{min}\{{\bm x}^{t_k*}\}_{k=1}^K$ and ${\bm u}^*=\mathop{max}\{{\bm x}^{t_k*}\}_{k=1}^K$. To deal with outliers in source datasets, this work also consider to add regularization parameter and slack variables to Eq.\ref{box}, which we will not introduce in details.

For ellipsoid search space, the parameter vector is ${\bm \theta}=({\bm A},{\bm b})$, where ${\bm A}\in\mathbbm{R}^{p\times p}$ is a symmetric positive definite matrix, and ${\bm b} \in \mathbbm{R}^p$ is an offset vector. The search space is designed as a hyperellipsoid $\hat{\mathcal{X}}({\bm \theta})=\{{\bm x}\in \mathbbm{R}^p\mid \|{\bm A}{\bm x}+{\bm b}\|_2\leqslant1\}$. Thus the constraint optimization is defined as:
\begin{equation}
    \min\limits_{{\bm A}\in\mathbbm{R}^{p\times p},{\bm A}\succ {\bm 0},{\bm b}\in\mathbbm{R}^p}\log\det({\bm A}^{-1})\ \text{such that for}\ k\in\{1,...,K\},\|{\bm A}{\bm x}^{t_k*}+{\bm b}\|_2\leqslant1.
\end{equation}

In practice, they apply a rejection sampling to guarantee uniform sampling. They first sample points uniformly in $p$-dimensional ball, and then they map the points into an ellipsoid. For more details, please refer to \cite{perrone2019learning}.\\

Also focused on search space design, \citet{li2022transfer} propose to leverage the information of similarities between different datasets to design a new search space for the problem, which has an uncertain space, different from \cite{perrone2019learning} using restricted geometrical shapes. Their main idea is that the more similar is between the source task and the target task, the more information can be leveraged from the source task to the target task. Based on that idea, they measure the similarities between source tasks and the target task also using the Kendall tau rank correlation coefficient~\cite{kendall1938new} as {Eq.\ref{KTRC}}, and then use that similarity to compute a fractile to choose points from each source task. Finally, a voting mechanism is used to combining the information from all source tasks to decide whether a point will be included in the new search space for the target task.
\section{Application Scenarios}
\label{sec_apps}
With the increasing use of Bayesian optimization in application scenarios, transfer learning-based methods can also be of use and help to reduce time and computational resources. 
Following we list some application scenarios that can take advantage of the progress of the transfer learning-based BO methods.

\subsection{AutoML Tuning}
Automated machine learning (AutoML) aims at tuning hyperparameters of machine learning models or choosing proper operations to construct task-specific neural architectures. 
In practice, when we need to tune a machine learning model, it is quite often that the model has been already tuned on various history datasets.
Transferring those knowledge saves the budget for re-training the models on the new task, especially when the evaluation cost is quite large, e.g., training deep neural networks or using huge datasets. 
Among the aforementioned literature, most methods~\cite{schilling2016scalable,wistuba2016two,feurer2018scalable,li2022transbo,li2022volcanomljournal} demonstrate powerful performance when transferring knowledge between the tuning knowledge of traditional machine learning models (e.g., Adaboost, SVM) on tabular datasets.
With the support of NASBench-201~\cite{dong2020bench}, recent work~\cite{li2022transbo,li2022transfer} shows that transfer learning also finds well-performed neural architectures quickly based on the tuning history of other datasets.

\subsection{DBMS Tuning}
Modern database management systems (DBMSs) have hundreds of configuration knobs that control their runtime behaviors (e.g., resource management, query optimizer).
Given a workload, DBMS tuning aims to  judiciously adjust the values of knobs  to optimize the system performance.
Tuning a DBMS is expensive, since it requires DBMS copies, computing resources, and the infrastructure to replay workloads and the tools to collect performance metrics~\cite{zhang2021facilitating}.
Therefore, transfer learning is adopted to leverage the tuning experience from the historical tasks and accelerate the tuning process of the new tasks.
Specifically, OtterTune\cite{DBLP:conf/sigmod/AkenPGZ17} and OnlineTune~\cite{DBLP:conf/sigmod/ZhangW0T0022} utilize the  observations from similar tuning tasks to train the target surrogate, which has shown to have better performance than tuning from scratch.
ResTune~\cite{DBLP:conf/sigmod/ZhangWCJT0Z021} adopts ensemble GPs (i.e., RGPE~\cite{feurer2018scalable}) to speedup the target tuning task.

\subsection{Computing Platform Tuning}
Big data computing platforms contain a huge number of parameters, for example, Hadoop~\cite{borthakur2007hadoop} and Spark~\cite{zaharia2010spark} each have over 200 parameters~\cite{bilal2017towards,kadirvel2012grey,singhal2018performance}. 
Meanwhile, these platforms have incredible scale and complexity, which requires system administrators to tune hundreds to thousands of nodes~\cite{herodotou2020survey}. 
To further accelerate the tuning process, transfer learning methods are taken into consideration. 
For example, \citet{wang2022industry} warm-start the tuning process using configurations in similar tasks.
Tuneful~\cite{DBLP:conf/kdd/FekryCPRH20} adopts Multi-Task Gaussian Processes~\cite{swersky2013multi} to utilize the most similar history task.





\section{Future Direction}
While transfer learning-based BO has gained huge progress in recent years, there are still some problems to be solved. In this section, we outline several promising prospective research directions.
\subsection{Evaluation Analysis}
Most previous work usually consider specific problems and use specific tasks to analyze their methods. 
They only compare themselves with a few baselines, e.g., they usually compare with classical methods like RGPE~\cite{feurer2018scalable}, TST~\cite{schilling2016scalable}, and multi-task BO~\cite{swersky2013multi}, or even simple BO without the transfer learning mechanism. 
Meanwhile, as different methods run tasks on the different application environments, it is hard to measure the performance gap between methods based on different experimental setups. 
Therefore, a comprehensive empirical analysis or a general benchmark is required to perform a fair comparison among transfer learning methods.

\subsection{Comprehensive Framework}
As introduced in Sec.~\ref{sec_overview}, the transfer learning for Bayesian optimization (TLBO) framework includes four main parts, the surrogate model, the acquisition function, the initial search points, and the search space. 
Most work consider only one aspect of this framework. 
We notice that the four parts are orthogonal to each other, which means that there is an opportunity to combine different methods into a comprehensive TLBO framework. 
To design such a framework, more challenges on how to combine those components can be discovered and addressed in future work.

\subsection{Generalized Transferable Information}
Previous TLBO methods mainly transfer the observations or information (e.g., meta-features) in history tasks. 
In practice, other types of knowledge can also be potential information to transfer, e.g., low-fidelity results~\cite{falkner2018bohb,li2021mfes} (evaluations with a proportion of time, epoches, data, etc.).
How to make use of other knowledge is also a promising direction to improve the performance of transfer learning.

\subsection{Combined Transfer Learning Method}
To accelerate the convergence of neural networks, various previous work~\cite{weiss2016survey,zhuang2020comprehensive} proposes to transfer trainable parameters from previous models to the new one.
However, the model still requires tuning hyperparameters to achieve strong performance.
While sharing the same spirit of TLBO methods that .
To further improve the model performance, it's interesting to discover how to perform a combined transfer of both trainable parameters and hyperparameters from previous tasks.

\section{Conclusion}
\label{sec_conclusion}
In this paper, we provided an in-depth review of the transfer learning methods for Bayesian optimization. 
First, based on ``what to transfer'' and ``how to transfer'', we systematically divide existing transfer learning works of Bayesian optimization into four categories:
initial point design-, search space design-, surrogate model-, and acquisition function-based approaches.
For each category, we presented the methodological design and technical descriptions in detail.
In addition, we investigated a general transfer learning framework for Bayesian optimization that considers all the four aspects, which can be a guidance for developing new approaches.
Finally, we showcased the potential application scenarios, where the transfer learning approaches for Bayesian optimization could work well.

\bibliographystyle{ACM-Reference-Format}
\bibliography{reference}


\end{document}